\documentclass[sigconf,authordraft,review=false]{acmart}
\usepackage[linesnumbered,ruled,lined]{algorithm2e}
\usepackage{xcolor}
\usepackage{array} 
\usepackage{amsfonts}
\usepackage{relsize}
\usepackage{graphicx}
\usepackage{subcaption}
\usepackage{multirow}
\usepackage{makecell}
\usepackage{tabularx}
\usepackage{pifont}
\usepackage{enumitem}
\AtBeginDocument{%
  \providecommand\BibTeX{{%
    \normalfont B\kern-0.5em{\scshape i\kern-0.25em b}\kern-0.8em\TeX}}}
\newcommand{\myhline}{\noalign{\global\arrayrulewidth0.3mm}\hline
                      \noalign{\global\arrayrulewidth0.3pt}}

\renewcommand\footnotetextcopyrightpermission[1]{}

\acmConference{Preprint}{Under Review}

\begin{document}

\title{SFedKD: Sequential Federated Learning with Discrepancy-Aware Multi-Teacher Knowledge Distillation}



\author{Haotian Xu}
\affiliation{%
  \institution{University of Science and Technology of China}
  \city{Hefei}
  \country{China}}
\email{xht2002@mail.ustc.edu.cn}

\author{Jinrui Zhou}
\affiliation{%
  \institution{University of Science and Technology of China}
  \city{Hefei}
  \country{China}}
\email{zzkevin@mail.ustc.edu.cn}

\author{Xichong Zhang}
\affiliation{%
  \institution{University of Science and Technology of China}
  \city{Hefei}
  \country{China}}
\email{yyu18@mail.ustc.edu.cn}

\author{Mingjun Xiao}
\affiliation{%
  \institution{University of Science and Technology of China}
  \city{Hefei}
  \country{China}}
\email{xiaomj@ustc.edu.cn}

\author{He Sun}
\affiliation{%
  \institution{University of Science and Technology of China}
  \city{Hefei}
  \country{China}}
\email{hesun@mail.ustc.edu.cn}

\author{Yin Xu}
\affiliation{%
  \institution{University of Science and Technology of China}
  \city{Hefei}
  \country{China}}
\email{yinxu@ustc.edu.cn}



\begin{abstract}
  Federated Learning (FL) is a distributed machine learning paradigm which coordinates multiple clients to collaboratively train a global model via a central server. Sequential Federated Learning (SFL) is a newly-emerging FL training framework where the global model is trained in a sequential manner across clients. Since SFL can provide strong convergence guarantees under data heterogeneity, it has attracted significant research attention in recent years. However, experiments show that SFL suffers from severe catastrophic forgetting in heterogeneous environments, meaning that the model tends to forget knowledge learned from previous clients. To address this issue, we propose an SFL framework with discrepancy-aware multi-teacher knowledge distillation, called SFedKD, which selects multiple models from the previous round to guide the current round of training. In SFedKD, we extend the single-teacher Decoupled Knowledge Distillation approach to our multi-teacher setting and assign distinct weights to teachers' target-class and non-target-class knowledge based on the class distributional discrepancy between teacher and student data. Through this fine-grained weighting strategy, SFedKD can enhance model training efficacy while mitigating catastrophic forgetting. Additionally, to prevent knowledge dilution, we eliminate redundant teachers for the knowledge distillation and formalize it as a variant of the maximum coverage problem. Based on the greedy strategy, we design a complementary-based teacher selection mechanism to ensure that the selected teachers achieve comprehensive knowledge space coverage while reducing communication and computational costs. Extensive experiments show that SFedKD effectively overcomes catastrophic forgetting in SFL and outperforms state-of-the-art FL methods.
\end{abstract}

\ccsdesc[500]{Computing methodologies~Distributed algorithms}

\keywords{Sequential Federated Learning, Catastrophic Forgetting, Multi-teacher Knowledge Distillation}



\maketitle

\begin{figure*}[t]
    \centering
    \begin{subfigure}{0.23\textwidth} 
        \centering
        \begin{minipage}{\textwidth}
            \centering
            \includegraphics[width=\textwidth]{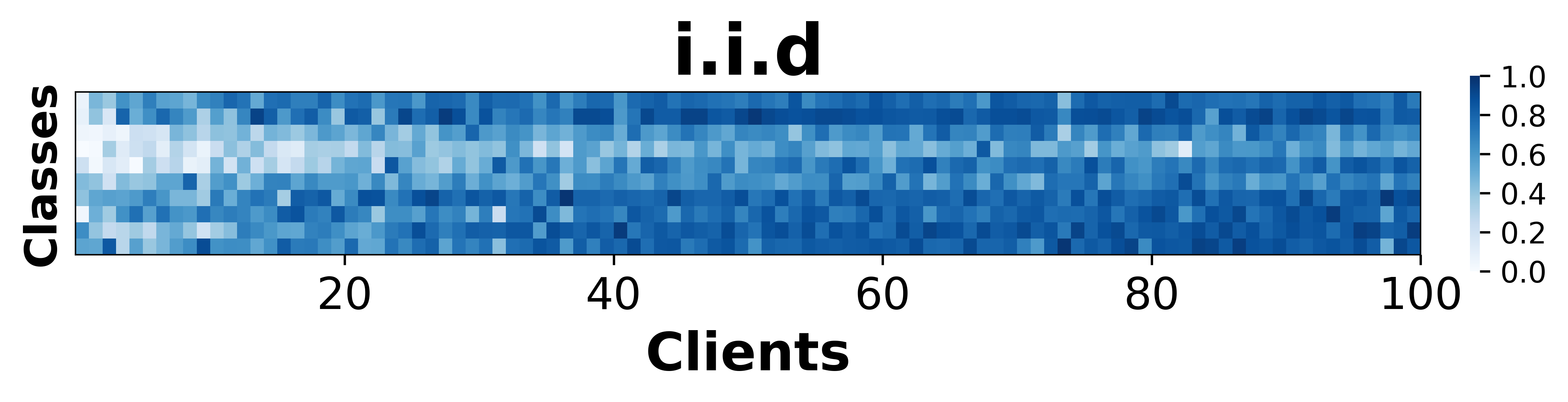} 
        \end{minipage}
        \vspace{0.8cm} 
        \begin{minipage}{\textwidth}
            \centering
            \includegraphics[width=\textwidth]{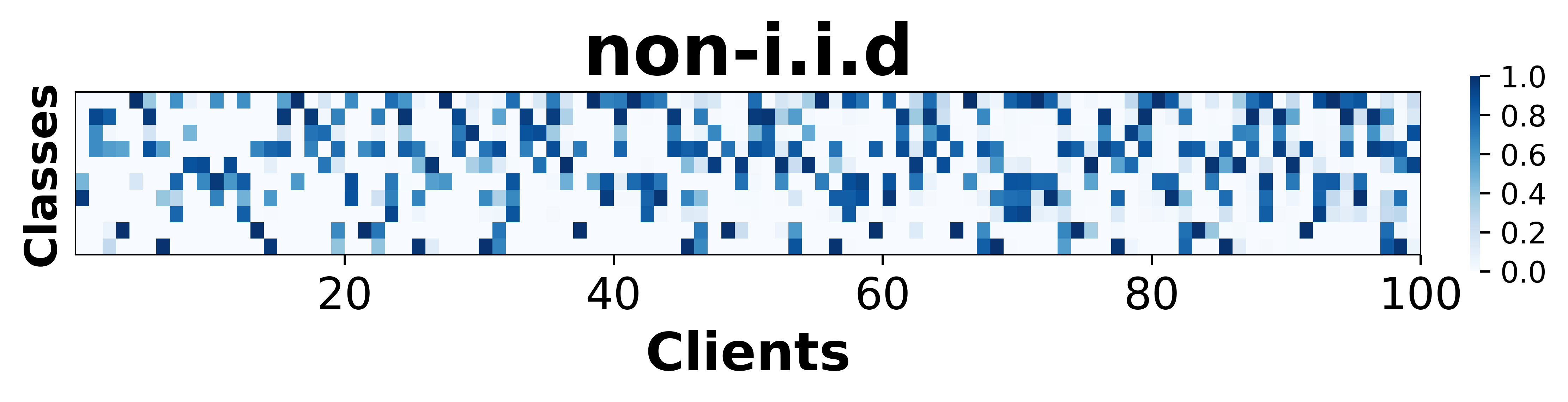} 
        \end{minipage}
        \caption{Class-wise Accuracy}
        \label{fig:class}
    \end{subfigure}
    \hspace{1mm} 
    \begin{subfigure}{0.24\textwidth}
        \centering
        \includegraphics[width=\textwidth]{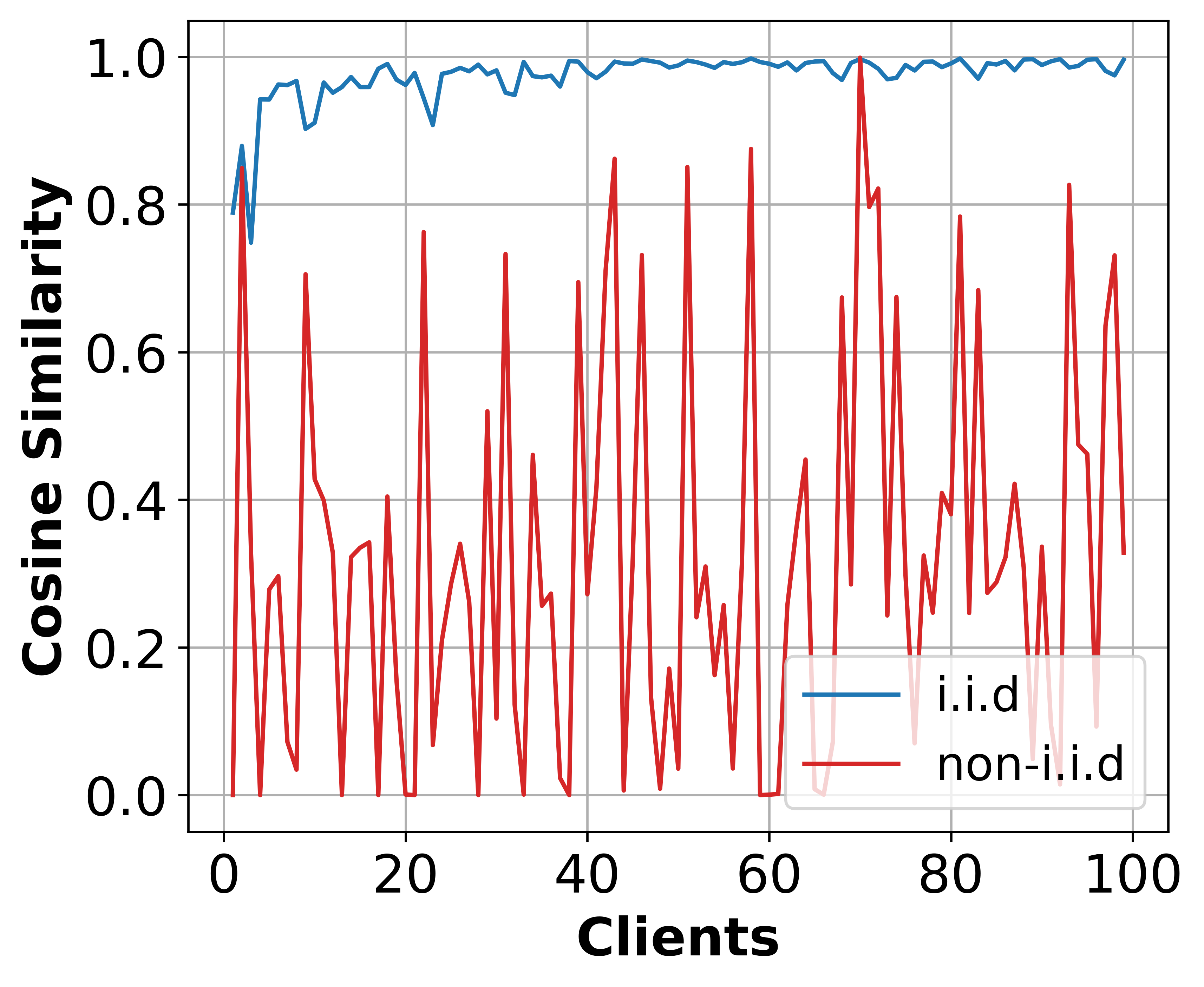}
        \caption{Model Consistency}
        \label{fig:consistency}
    \end{subfigure}
    \hspace{1mm}
    \begin{subfigure}{0.24\textwidth}
        \centering
        \includegraphics[width=\textwidth]{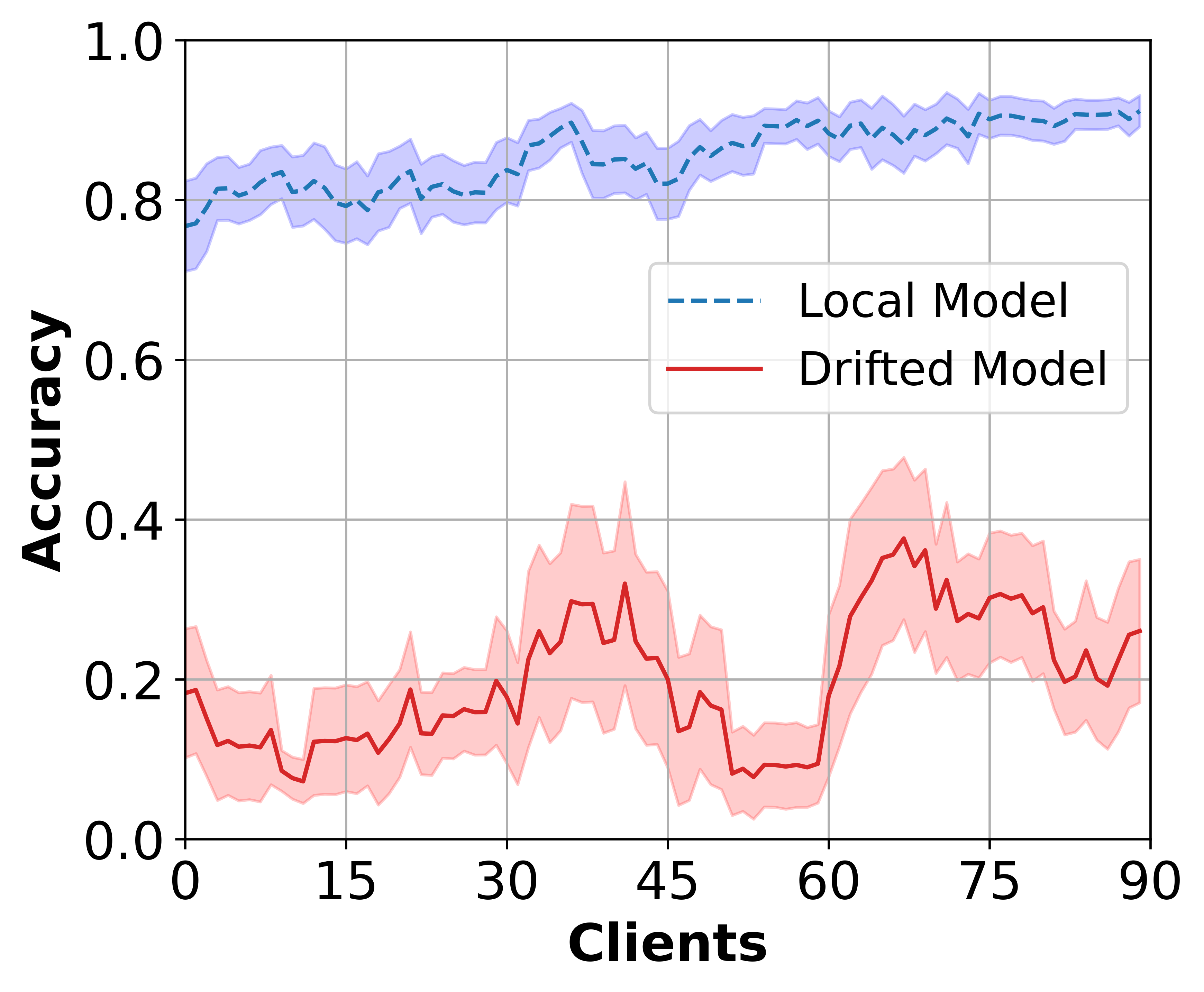}
        \caption{Model Drift}
        \label{fig:drift}
    \end{subfigure}
    \hspace{1mm}
    \begin{subfigure}{0.24\textwidth}
        \centering
        \includegraphics[width=\textwidth]{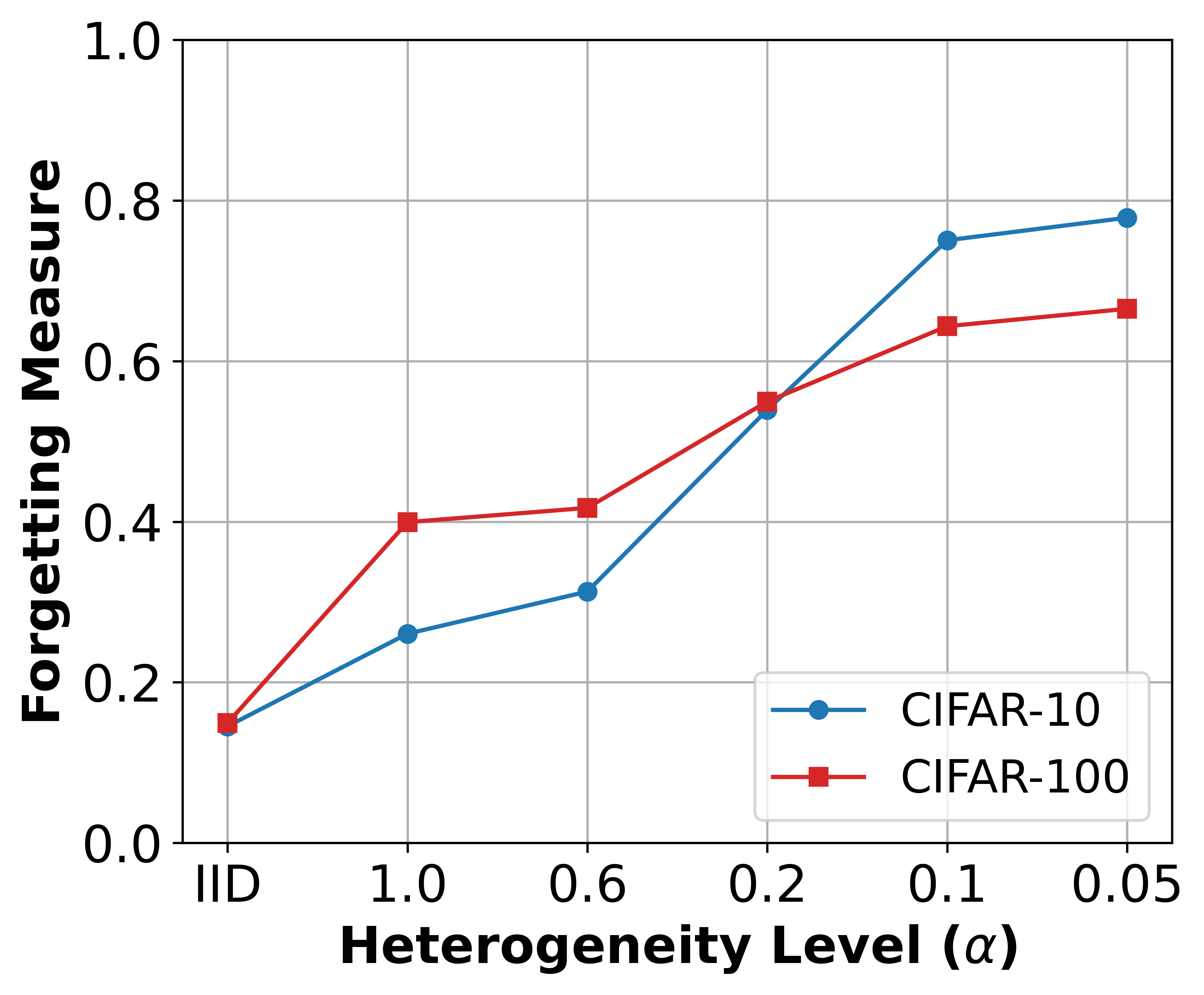}
        \caption{Forgetting Measure}
        \label{fig:measure}
    \end{subfigure}
    \vspace{-5pt}
    \caption{Forgetting analysis on the model trained on a sequence comprising 100 clients under different scenarios. (a) Class-wise test accuracy after training by each client in i.i.d. and non-i.i.d. scenarios on CIFAR-10. (b) Cosine similarity between the class-wise test accuracy vectors of the models trained by the consecutive clients in i.i.d. and non-i.i.d. scenarios. (c) Accuracy of the local model trained on the current client and the drifted model trained on the next client when evaluated on the local dataset of the current client on CIFAR-10. (d) Forgetting Measure under different heterogeneity levels on CIFAR-10 and CIFAR-100.}
    \label{fig:fogetting}
\end{figure*}

\section{Introduction}
Federated Learning (FL)~\cite{mcmahan2017communication} is a distributed machine learning paradigm which enables multiple clients to jointly train a global model under the coordination of a central server without sharing their local data with each other, which provides significant advantages in dealing with the well-known ``isolated data islands'' issue. Many works have been conducted to investigate various challenges in FL, such as statistical heterogeneity~\cite{li2019convergence,lu2023federated,qi2023cross,lee2022preservation,an2023federated}, communication~\cite{liu2024fedbcgd,li2024masked,wang2022communication,rothchild2020fetchsgd,hamer2020fedboost}, and cross-modal~\cite{li2024cross,fu2024fedcafe,li2023prototype,xiong2023client,zuo2024privacy}. According to the training mode, FL can be roughly categorized into Parallel FL (PFL) and Sequential FL (SFL). In the PFL framework, clients train local models in parallel and upload them to a central server to be aggregated into a global model, such as FedAvg~\cite{mcmahan2017communication}. In the SFL framework, models are trained in a sequential manner across clients~\cite{chang2018distributed}. Although SFL might lead to a longer single-round training time, it can converge in fewer rounds than PFL, demonstrating better convergence guarantees, when addressing heterogeneous data~\cite{li2024convergence}, especially for small-scale datasets~\cite{kamp2021federated}. Thus, numerous related works have been proposed in recent years~\cite{chen2023metafed, wang2024one,zhou2025psfl}.



Although SFL has demonstrated excellent performance in various scenarios, it still faces a critical challenge. The model performance exhibits significant \textbf{catastrophic forgetting}~\cite{mccloskey1989catastrophic} on heterogeneous data, which stems from the model overfitting to the local data of the current client while forgetting knowledge from previous clients. We conduct a series of SFL experiments to validate the issues, as shown in Fig.~\ref{fig:fogetting} (the detailed settings are described in Sect. 4). For instance, Fig.~\ref{fig:class} and Fig.~\ref{fig:consistency} illustrate the model's class-wise test accuracy when trained sequentially by clients in the i.i.d. and non-i.i.d. settings. The non-i.i.d. setting exhibits significant fluctuations, implying the existence of catastrophic forgetting. Fig.~\ref{fig:drift} demonstrates that while the local model achieves optimal performance on the current client's dataset, this knowledge is rapidly forgotten during subsequent training on the next client, leading to a drifted model. Fig.~\ref{fig:measure} presents that as the heterogeneity level increases, the value of Forgetting Measure (a metric defined in~\cite{chaudhry2018riemannian}) increases, which indicates a higher degree of forgetting. However, existing SFL research has scarcely investigated the catastrophic forgetting issue. So far, only the CWC framework~\cite{song2024cyclical} has been proposed to address this issue, which utilizes an importance matrix to stabilize critical parameters, thereby preserving prior knowledge and mitigating catastrophic forgetting. Nevertheless, the importance matrix may conflict in different clients, making it hard to optimize the model and leading to poor performance on new clients~\cite{zhou2021co, zhou2024class}.



In this paper, we introduce the multi-teacher Knowledge Distillation (KD) technique into SFL to address the catastrophic forgetting issue. First, we propose a discrepancy-aware multi-teacher KD method, where multiple models trained in the previous round are selected to serve as teacher models (for abbr., teachers) to guide the current round of training. Since the predictive performance of teacher models on different classes is determined by their corresponding clients' local class distributions, each teacher possesses distinct class-specific knowledge valuable for distillation. To maintain consistent performance across all classes and avoid local optima caused by imbalanced data, the student model must engage in personalized knowledge acquisition from diverse teachers. To simultaneously enhance model training efficacy and mitigate catastrophic forgetting, we extend the single-teacher Decoupled Knowledge Distillation (DKD)~\cite{zhao2022decoupled} approach to our multi-teacher setting and assign distinct weights to teachers’ target-class and non-target-class knowledge based on the class distributional discrepancies between their corresponding clients' and the current client's local data. Second, we design a complementary-based teacher selection mechanism to select appropriate teachers to maximize knowledge space coverage, which is formalized as a variant of the maximum coverage problem to be solved. This mechanism can address the knowledge dilution caused by redundant teachers~\cite{yang2020model} while reducing communication and computational costs. Based on the two designs, we propose a \underline{S}equential \underline{Fed}erated learning framework with discrepancy-aware multi-teacher \underline{K}nowledge \underline{D}istillation (SFedKD). The main contributions are summarized as follows:

\begin{itemize}
    \item We investigate the catastrophic forgetting issue in SFL through serious experiments and propose a novel SFL framework, i.e., SFedKD, to solve this issue by introducing multi-teacher KD. To the best of our knowledge, it is the first work to utilize multi-teacher KD to tackle the catastrophic forgetting in SFL.
    \item We propose a discrepancy-aware multi-teacher KD method for SFedKD by extending the single-teacher DKD approach to our multi-teacher setting and assigning distinct weights to teachers’ target-class and non-target-class knowledge based on local class distributional discrepancies between teacher clients and student client, which can enhance model training efficacy while mitigating catastrophic forgetting.
    \item We design a complementary-based teacher selection mechanism that preserves the diversity of knowledge within the teachers while reducing communication and computational costs, which significantly prevents knowledge dilution.
    \item We conduct extensive experiments on five datasets by varying degrees of heterogeneity. The results demonstrate that our method achieves superior performance compared with existing PFL and SFL methods.
    
\end{itemize}

\begin{figure*}[t]
    \centering
     \includegraphics[width=0.95\linewidth]{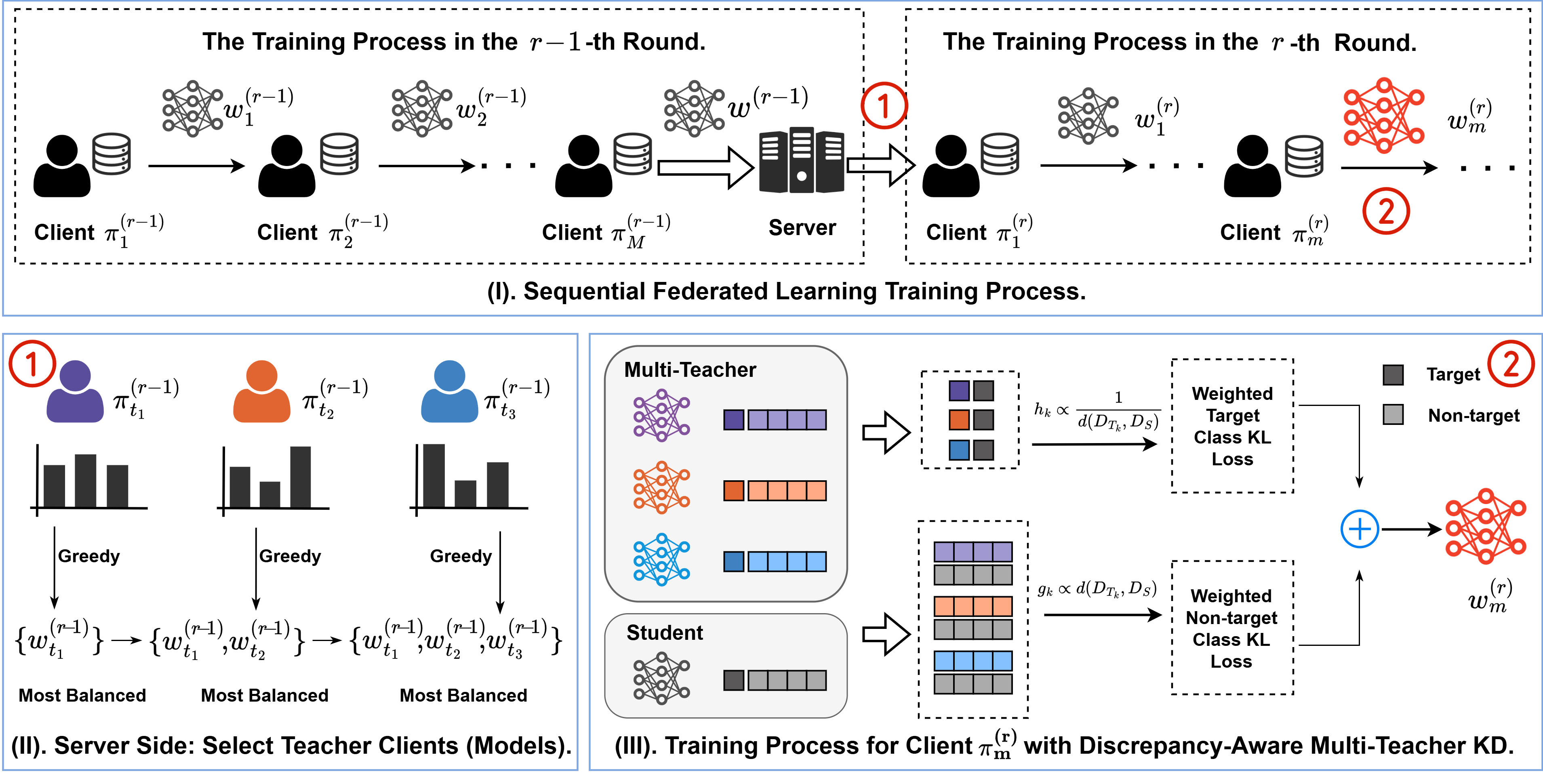}
     \caption{ The overview of our proposed SFedKD. The model is trained in a sequential manner across clients (I). At the beginning of the $r$-th round, the server applies a greedy teacher selection algorithm to select teacher clients and collect corresponding teacher models from the $(r-1)$-th round (II). The teacher models guide the training process in the $r$-th round utilizing our proposed discrepancy-aware multi-teacher KD (III). }
     \label{fig:overview}
\end{figure*}

\section{Related Work}

\subsection{Sequential Federated Learning}
SFL trains models sequentially across clients~\cite{chang2018distributed}, in contrast to PFL, where clients train models in parallel~\cite{mcmahan2017communication}. As a new paradigm of FL, SFL has been proven to achieve better convergence guarantees than PFL in terms of training rounds on heterogeneous data~\cite{li2024convergence}. For example,  FedDC~\cite{kamp2021federated} enhances model performance on small datasets via a combination of model permutations across clients and aggregation. MetaFed~\cite{chen2023metafed} is an SFL framework with cyclic knowledge distillation for personalized healthcare. FedELMY~\cite{wang2024one} is a one-shot SFL framework that enhances global model performance under low communication costs by explicitly promoting model diversity during local training. PSFL~\cite{zhou2025psfl} is a hybrid framework that combines parallel and sequential training modes to accelerate convergence on heterogeneous data. However, all of these works have not investigated the catastrophic forgetting issue induced by data heterogeneity in SFL. So far, only the CWC~\cite{song2024cyclical} utilizes the importance matrix to regularize the local optimization process, yet it risks matrix conflicts across clients. In contrast, our SFedKD framework employs discrepancy-aware multi-teacher KD to enable personalized assimilation of previous model knowledge, achieving superior catastrophic forgetting mitigation.



\subsection{Catastrophic Forgetting}
Catastrophic Forgetting~\cite{mccloskey1989catastrophic, french1999catastrophic} is widely recognized as a fundamental challenge in neural networks. This phenomenon often arises in scenarios involving multiple sequential tasks. When trained on new tasks, neural networks often exhibit significant performance degradation on previously learned tasks, effectively losing acquired knowledge. Different approaches have been proposed to mitigate catastrophic forgetting. In the replay-based strategy~\cite{aljundi2019gradient, shin2017continual, rebuffi2017icarl, chaudhry2019tiny, rolnick2019experience}, a network can overcome catastrophic forgetting by using a replay buffer and revisiting former exemplars. Catastrophic forgetting is also addressed by parameter regularization methods~\cite{zenke2017continual, lee2017overcoming, kirkpatrick2017overcoming, farajtabar2020orthogonal} that estimate the importance of network parameters and constrain updates to critical parameters to preserve previously learned knowledge. There are also knowledge distillation methods~\cite{li2017learning, hou2018lifelong, wu2019large, lee2019overcoming, kang2022class, guo2024continual, wen2024class} that build distillation relationship between the teacher model trained on previous tasks and the student model trained on the current task to transfer knowledge from the former to the latter, among which~\cite{guo2024continual, wen2024class} have explored multi-teacher knowledge distillation approaches. In SFL with heterogeneous clients, catastrophic forgetting occurs because the model over-adapts to the current client's local data, discarding knowledge from previous clients. We propose a discrepancy-aware multi-teacher knowledge distillation method to address it, unlike these works. 


\subsection{Multi-Teacher Knowledge Distillation}
Unlike traditional Knowledge Distillation (KD)~\cite{hinton2015distilling}, where the student learns from a single teacher, multi-teacher KD enables the student to acquire more comprehensive knowledge by aggregating information from multiple teachers. The student distills knowledge from either the aggregated softened output logits of all teachers (class-level knowledge)~\cite{mirzadeh2020improved, zhu2018knowledge, tarvainen2017mean, you2017learning} or the ensemble of their feature representations aligned via similarity matrices (structural knowledge)~\cite{wu2019distilled, liu2019knowledge, park2019feed}. In previous works, teacher weights are often assigned equally~\cite{tarvainen2017mean, mirzadeh2020improved, you2017learning, papernot2016semi}. However, subsequent research recognizes the varying quality and importance of different teachers and designs weights based on factors such as entropy~\cite{kwon2020adaptive}, latent factor~\cite{liu2020adaptive}, and prediction confidence~\cite{zhang2022confidence}. Notably, \cite{cheng2024mkd} proposes a decoupled multi-teacher KD based on entropy. Different from them, we extend the single-teacher Decoupled Knowledge Distillation~\cite{zhao2022decoupled} approach to our multi-teacher setting and assign distinct weights to the teachers’ target-class and non-target-class knowledge based on the class distributional discrepancies between the teacher clients' and the student client's local data.


\section{METHOD}
\subsection{Problem and Framework}

We consider a typical FL system, consisting of a server and a client set $\mathcal{N} = \{1, 2, \cdots, N \}$. Each client $n \in \mathcal{N}$ holds its local dataset $\mathcal{B}_n=\{\xi_i^n\}_{i=1}^{|\mathcal{B}_n|}$ with the size of $|\mathcal{B}_n|$, where $\xi_i^n$ denotes a data sample from $\mathcal{B}_n$. The local class distribution of $n$-th client’s dataset is defined as $\mathbf{D}_n\in \mathbb{R}^C$, where the $c$-th element $\mathbf{D}_{n,c}=\frac{|\{(\mathbf{x}_i,y_i)|y_i=c\}|}{\mathcal{B}_n}$, $c\in \{1,2,\cdots,C\}$ is the data proportion of the $c$-th class. We define $\mathcal{L}(\mathbf{w}; \xi)$ as a loss function to evaluate a machine learning model $\mathbf{w}$ which performs on the data sample $\xi$. The optimization objective is formulated as:
\begin{equation}
    \mathop{\min}_{\mathbf{w}}\sum_{n=1}^N \mathbb{E}_{\xi\sim \mathcal{B}_n} [\mathcal{L}(\mathbf{w};\xi)].
\end{equation}

In this work, we propose an SFL framework SFedKD, along with a discrepancy-aware multi-teacher KD approach to address the catastrophic forgetting problem in traditional SFL and a teacher selection mechanism to filter out redundant teacher models for efficient distillation. The overview of our framework is demonstrated in Fig.~\ref{fig:overview}. We consider that the whole SFL process consists of $R$ rounds, and let $r \in \{1,2,\cdots,R\} $ indicate the rounds. The SFedKD framework works as follows.

At the beginning of each round $r$, the server randomly selects a sequence of clients from the set $\mathcal{N}$ through sampling without replacement, denote as $\Pi^{(r)}=\{\pi_1^{(r)},\pi_2^{(r)},\cdots,\pi_M^{(r)}\}$. Here, $\pi_m^{(r)}$ is the index of the $m$-th sampled client in the $r$-th round and $M$ clients are sampled in each round. The $m$-th trained local model in the sequence during round $r$ is denoted by $\mathbf{w}_m^{(r)}$. The server distributes the global model $\mathbf{w}^{(r-1)}$ obtained at the end of the previous round to the first client in the sequence to start the current round of training by setting $\mathbf{w}_0^{(r)}=\mathbf{w}^{(r-1)}$ (Fig.~\ref{fig:overview} (I)). Meanwhile, the server will select a subset of teacher clients $\mathcal{T}^{(r)}=\{\pi_{t_1}^{(r-1)}, \pi_{t_2}^{(r-1)},\cdots,\pi_{t_K}^{(r-1)}\} \subseteq \Pi^{(r-1)}$ from the previous round through the teacher selection mechanism and collect their training models $\mathcal{W}^{(r)}=\{w_{t_1}^{(r-1)}, w_{t_2}^{(r-1)},\cdots,w_{t_K}^{(r-1)}\}$ as teacher models to guide this round of training (Fig.~\ref{fig:overview} (II)), where $K$ is the number of  teachers. Here, we call the client that trains a teacher model the corresponding teacher client.

During the $r$-th round, the selected clients in the sequence $\Pi^{(r)}$ train the model one-by-one. Concretely, the $m$-th client $\pi_m^{(r)}$ receives the model trained by the previous client $\pi_{m-1}^{(r)}$. Then, it trains the model to learn from its local data while utilizing the teacher models in the subset $\mathcal{W}^{(r)}$ by performing the discrepancy-aware multi-teacher KD method based on the discrepancies between $\{\mathbf{D}_{\pi_{t_1}^{(r-1)}},\cdots,\mathbf{D}_{\pi_{t_k}^{(r-1)}},\cdots,\mathbf{D}_{\pi_{t_K}^{(r-1)}}\}$ and $\mathbf{D}_{\pi_m^{(r)}}$ (Fig.~\ref{fig:overview} (III)), where $\mathbf{D}_{\pi_{t_k}^{(r-1)}}$ is the $k$-th teacher client's local class distribution. The training performs Stochastic Gradient Descent~\cite{bottou2010large} as follows:

    \begin{equation}
    \mathbf{w}_{m}^{(r)} = \mathbf{w}_{m-1}^{(r)}-\eta\nabla \mathcal{L}(\mathbf{w}^{(r)}_{m-1}, \boldsymbol{\xi}_{\pi_m^{(r)}})
    ,
    \end{equation}
where $\boldsymbol{\xi}_{\pi_m^{(r)}}$ is the mini-batch sampled from $\mathcal{B}_{\pi_m^{(r)}}$, $\eta$ is the learning rate. After training on the client $\pi_m^{(r)}$ is completed, the model $\mathbf{w}_m^{(r)}$ is passed to the server and the next client $\pi_{m+1}^{(r)}$ in the sequence.

\subsection{Discrepancy-Aware Multi-Teacher KD}


In this subsection, we propose the discrepancy-aware multi-teacher KD method for the SFedKD framework. Note that multiple teacher models will be selected to guide the current round of training, and these teacher models contribute to each client's training differently. Although traditional multi-teacher KD methods~\cite{kwon2020adaptive, zhang2022confidence} can characterize personalized teacher contributions, they fail to effectively balance training effectiveness and catastrophic forgetting mitigation due to their rough weight strategies. Therefore, we extend the single-teacher Decoupled Knowledge Distillation~\cite{zhao2022decoupled} approach to our multi-teacher setting. Moreover, we utilize its two decoupled components, Non-target Class Knowledge Distillation (NCKD) and Target Class Knowledge Distillation (TCKD), and assign distinct weights to them respectively based on the class distributional discrepancies between the teacher clients' and the student client's local data. Through this fine-grained weighting strategy, the former component specifically targets catastrophic forgetting mitigation, while the latter enhances model training efficacy, collectively establishing an optimal balance to improve overall training performance.



First, we extend the components NCKD and TCKD to our multi-teacher setting. NCKD represents the weighted KL-Divergence between the teacher and student models' softmax probability vectors over all non-target classes, formulated as:
\begin{equation}
\begin{gathered}
{\mathcal{L}}_{NCKD}=\sum_{k=1}^K g_k\sum_{c=1\,, \underline{\mathbf{c \neq t}}}^{C}{\tilde{p}_{c}^{T_k}}\log\left[\frac{{\tilde{p}_{c}^{T_k}}}{{\tilde{p}_{c}^{S}}}\right]
, \; \text{where}\\
\tilde{p}_{c}^{T_k} =\frac { \exp{({z}_{c}^{T_k}/ \tau)}}{\sum_{\underline{\mathbf{\tilde{c} \neq t}}}^{C}\exp{({z}_{\tilde{c}}^{T_k}/\tau)}}
, \:
\tilde{p}_{c}^{S} =\frac { \exp{({z}_{c}^{S}/ \tau)}}{\sum_{\underline{\mathbf{\tilde{c} \neq t}}}^{C}\exp{({z}_{\tilde{c}}^{S}/\tau)}}
, \:\quad \forall c \neq t,
\end{gathered}
\label{NCKD}
\end{equation}
where $\mathbf{z}=[z_1,\cdots,z_t,\cdots,z_C]$ is the logits output of training sample, $t$ is the target class, $\tau$ is temperature, $T_k$ is the $k$-th teacher, $g_k$ represents the corresponding weight, $K$ is the number of teachers, and $S$ denotes the student. TCKD is the weighted KL-Divergence between the binary probabilities of the target class and all the other non-target classes from teacher and student models, formulated as:
\begin{equation}
\begin{gathered}
{\mathcal{L}}_{TCKD}=\sum_{k=1}^K h_k({p_{t}^{T_k}}\log\left[\frac{{p_{t}^{T_k}}}{{p_{t}^{S}}}\right] + {p_{\tilde{t}}^{T_k}}\log\left[\frac{{p_{\tilde{t}}^{T_k}}}{{p_{\tilde{t}}^{S}}}\right])
, \; \text{where}\\
p_{t}^{T_k} =\frac { \exp{({z}_{t}^{T_k}/ \tau)}}{\sum_{\mathbf{c=1}}^{C}\exp{({z}_{c}^{T_k}/\tau)}}
, \:
p_{t}^{S} =\frac { \exp{({z}_{t}^{S}/ \tau)}}{\sum_{\mathbf{c=1}}^{C}\exp{({z}_{c}^{S}/\tau)}}\\
p_{\tilde{t}}^{T_k} =\frac { {\sum_{\underline{\mathbf{\tilde{c} \neq t}}}^{C}\exp{({z}_{\tilde{c}}^{T_k}/ \tau)}}} {\sum_{\mathbf{c=1}}^{C}\exp{({z}_{c}^{T_k}/\tau)}}
, \:
p_{\tilde{t}}^{S} =\frac { {\sum_{\underline{\mathbf{\tilde{c} \neq t}}}^{C}\exp{({z}_{\tilde{c}}^{S}/ \tau)}}} {\sum_{\mathbf{c=1}}^{C}\exp{({z}_{c}^{S}/\tau)}}.
\end{gathered}
\label{TCKD}
\end{equation}
Here, $h_k$ is the TCKD's weight of teacher $T_k$.


Second, we design personalized weight values to $g_k$ in Eq.~(\ref{NCKD}) and $h_k$ in Eq.~(\ref{TCKD}) based on the discrepancy between the local class distribution of each teacher client and the student client. For conciseness, we use $\mathbf{D}^{T_k}$ and $\mathbf{D}^S$ to represent the local class distribution of the $k$-th teacher client and the student client where the model is performing distillation, respectively. Due to the heterogeneity of data distributions across clients, each teacher model contains different knowledge, which is determined by the local class distribution of its training client. Meanwhile, the student client's local data follows an uneven distribution, which induces imbalanced teacher contributions because of their divergent performance on this data. As a result, the weights of each teacher are closely related to the discrepancy between its corresponding class distribution $\mathbf{D}^{T_k}$ and the student client’s class distribution $\mathbf{D}^S$. For NCKD, the greater the discrepancy between $\mathbf{D}^{T_k}$ and $\mathbf{D}^S$, the higher the weight of $T_k$, since it contains more class knowledge that the student client cannot directly acquire. Conversely, for TCKD, the more similar $\mathbf{D}^{T_k}$ and $\mathbf{D}^S$ are, the higher the weight of $T_k$, as its class knowledge is more relevant to what the student client needs to learn. To effectively aggregate the knowledge of multiple teachers, we define $g_k$ and $h_k$ as follows:
\begin{equation}
    g_k = \frac{d(\mathbf{D}^{T_k}, \mathbf{D}^S)}{\sum_{j}d(\mathbf{D}^{T_j}, \mathbf{D}^S)}
    ,\:
    h_k = \frac{\frac 1 {d(\mathbf{D}^{T_k}, \mathbf{D}^S)+\epsilon}}{\sum_{j}\frac 1 {d(\mathbf{D}^{T_j}, \mathbf{D}^S)+\epsilon}}
    ,
    \label{weight}
\end{equation}
where $\epsilon=1e-4$ is the smoothing constant and $d(\cdot)$ is a pre-defined metric function evaluating the difference between two distributions (e.g., L1 distance or KL-Divergence). For example, when applying L1 distance, it is instantiated as $d(\mathbf{D}^T, \mathbf{D}^S)=\sum_{c=1}^C|\mathbf{D}^{T,c}-\mathbf{D}^{S,c}|$. This weighting design enables the model to simultaneously achieve effective learning on current data and alleviate forgetting of prior knowledge, thereby maintaining model consistency and mitigating model shift throughout the training process.


\subsection{Teacher Selection Mechanism}
In this subsection, we propose the teacher selection mechanism for the SFedKD framework. Since teacher clients may share overlapping class distributions, some teachers can produce highly similar knowledge representations. These redundant teachers can dilute the effective knowledge, ultimately degrading the distillation performance. Moreover, increasing the number of teacher models imposes significant system overhead, including greater storage requirements, higher communication costs, and additional logit computations, which collectively degrade efficiency. Therefore, we need to select appropriate teachers that simultaneously maximize knowledge space coverage and minimize model redundancy.

We already know that the class knowledge contained in the teacher models is closely related to the local class distribution of their corresponding teacher clients, which implies that comprehensive knowledge space coverage requires the aggregated data distribution across teacher clients to approach uniformity. Then, the optimal teacher selection problem can be formalized as follows:
\begin{equation}
    \underset{\{\pi_{t_1}, \pi_{t_2}, \ldots, \pi_{t_K}\} \subseteq \{\pi_1, \pi_2, \ldots, \pi_M\}}{\arg\min} d (\sum_{k=1}^K \mathbf{D}_{\pi_{t_k}}, \mathbf{U} )
    ,
\end{equation}
where $K$ is the number of teachers, $\mathbf{U}$ is the uniform distribution, and $d(\cdot)$ is a distribution difference metric, which decreases as the distributions become more similar. Consider a special case of the above problem, where each client's local dataset contains only one class. This is a well-known classic maximum coverage problem, which is NP-hard~\cite{khuller1999budgeted}. Actually, the problem can be seen as a variant of the maximum coverage problem, and thus it is also NP-hard. There is no polynomial-time optimal algorithm for this problem, so we develop a greedy teacher selection algorithm to solve it, which is depicted in Algorithm~\ref{greedy}. We select the optimal client $\pi_{{t}_{k}}$ that minimizes the discrepancy between the aggregated class distribution $\mathbf{D}_{agg}$ and uniform distribution $\mathbf{U}$ in the $k$-th iteration until reaching the target teacher number $K$ (Lines 1-6). 

Through the above mechanism, the selected teacher clients exhibit a complementary relationship, as their aggregated class distribution approximates a uniform distribution, ensuring comprehensive coverage of the class knowledge space. In addition, our selection strategy actively prunes redundant teachers, thereby preventing knowledge dilution while reducing communication and computation costs.

\begin{algorithm}
    \caption{The Teacher Selection Algorithm}
    \label{greedy}
    \KwIn{clients $\Pi=\{\pi_1,\pi_2,\cdots,\pi_{M}\}$, number of sample teachers $K$, local class distribution vectors of the clients $\mathcal{D} = \{\mathbf{D}_{\pi_1},\mathbf{D}_{\pi_2},\cdots,\mathbf{D}_{\pi_{M}}\}$ }
    \KwOut{The teacher client set $\mathcal{T}$}
    \textbf{Initialization:} Initialize $\mathcal{T}=\emptyset$ and $\mathbf{D}_{agg}=\emptyset$.

    
    

    \While{$|\mathcal{T}| < K$}
    {
        \textcolor{blue}{// Select the client that makes the aggregation class distribution closest to a uniform distribution}
    
         ${t}_{|\mathcal{T}|+1}=\arg\min_{{t}}d(\mathbf{D}_{agg}+\mathbf{D}_{{\pi_t}}, \mathbf{U})$ for ${\pi_t}\in \Pi\setminus \mathcal{T}$
        
         $\mathcal{T} \leftarrow \mathcal{T}\cup \pi_{{t}_{min}}$, $\mathbf{D}_{agg}\leftarrow \mathbf{D}_{agg}+\mathbf{D}_{\pi_{{t}_{|\mathcal{T}|+1}}}$
    }
    
\end{algorithm}

\begin{algorithm}
    \caption{Our proposed SFedKD}
    \label{method}
    \KwIn{clients $\mathcal{N} = \{1,2,\cdots,N\}$, number of selected clients per round $M$, learning rate $\eta$, training round $R$}
    \KwOut{The final model $w_{final}$}
    
    \textbf{Initialization:} Initialize $w_0^{(1)}$ as the model for the first client in the first round.
    
        \For{round $r=1:R$}{
        
        The server samples a sequence $\pi_1^{(r)}, \pi_2^{(r)},\cdots,\pi_M^{(r)}$ of $\mathcal{N}$.

        \textcolor{blue}{// Execute the teacher selection mechanism}
        
        \If {$r\neq1$}{ The server runs Algorithm~\ref{greedy} to get teacher clients $\mathcal{T}$ and collects corresponding teacher models $\mathcal{W}$}
        
        \For{$m=1,2,\cdots,M$ {\bf{in sequence}}}{
    
            Optimize the total loss function $\mathcal{L}$ in Eq.~\ref{overall}
            $w_{m}^{(r)} \leftarrow w_{m-1}^{(r)} - \eta \nabla\mathcal{L}(w_{m-1}^{(r)}$)
		}
        $w^{(r)} \leftarrow w^{(r)}_{M}$, $w_0^{(r+1)} \leftarrow w^{(r)}$
        }

        \textcolor{blue}{// For the final client, output model $w_{final}$}
        
        $w_{final} \leftarrow w^{(R)}$
\end{algorithm}

\subsection{Overall Loss Function and Algorithm}
In summary, our loss function can be summarized as:
\begin{equation}
    \mathcal{L}=\mathcal{L}_{CE}+\gamma{\mathcal{L}}_{NCKD}+\beta{\mathcal{L}}_{TCKD}
    ,
    \label{overall}
\end{equation}
where $\mathcal{L}_{CE}$ is the cross-entropy loss function that measures the difference
between the prediction and the ground truth labels, $\gamma$ and $\beta$ denote the corresponding trade-off coefficients, $\mathcal{L}_{NCKD}$ and $\mathcal{L}_{TCKD}$ are respectively defined in Eq.~(\ref{NCKD}) and Eq.~(\ref{TCKD}).

For a better understanding of our method, we show the pseudocode of our framework SFedKD in Algorithm~\ref{method}. At the beginning of each round, the server samples a client sequence and executes the teacher selection mechanism to get teachers (Lines 3-7). Then, the sampled clients train the model according to the overall loss function defined in Eq.~\ref{overall} in sequence (Lines 8-11). At last, the final client outputs the ultimate global model (Lines 13-14).

\begin{table*}[t]
    \centering
    \renewcommand{\arraystretch}{1.1}
    \caption{Test accuracy (\%, mean$\pm$std on 5 trials) comparison of our SFedKD method to other baselines on several heterogeneous settings and datasets. FedSeq, CWC and our SFedKD are SFL methods, while other baselines are all PFL methods. Experiments show that SFedKD consistently outperforms these state-of-the-art methods.}
\begin{tabular}{c|cc|cc|cc|c|c}
\myhline

\multirow{2}{*}{Method} & \multicolumn{2}{c|}{Fashion-MNIST} & \multicolumn{2}{c|}{CIFAR-10} & \multicolumn{2}{c|}{CINIC-10} & \multicolumn{1}{c|}{CIFAR-100} & \multicolumn{1}{c}{HAM10000} \\
 & Exdir(2,0.5)   & Exdir(2,10.0)   & Exdir(2,0.5)   & Exdir(2,10.0) & Exdir(2,0.5)   & Exdir(2,10.0) &  Exdir(20,1.0) & Exdir(2,1.0) \\
\hline
FedAvg &   76.00$\pm$2.37    & 82.91$\pm$0.71 &   32.48$\pm$2.01    & 54.19$\pm$1.27 &  28.90$\pm$1.16     & 43.93$\pm$0.80 &  33.11$\pm$0.22     &     48.23$\pm$1.06   \\
FedProx &    76.13$\pm$2.16   & 82.89$\pm$0.78 &    32.48$\pm$2.02   & 53.69$\pm$1.25 &    29.25$\pm$1.02   & 44.00$\pm$0.62 &  33.00$\pm$0.13     &    48.08$\pm$0.92    \\
FedCurv &   75.73$\pm$2.25    & 82.95$\pm$0.83 &     32.34$\pm$1.85 & 54.15$\pm$1.04 &    28.81$\pm$1.18  & 43.91$\pm$0.74 &  32.99$\pm$0.33     &    47.94$\pm$1.06     \\
FedNTD & 76.18$\pm$3.05 & 82.62$\pm$0.86 & 32.27$\pm$1.40 & 53.87$\pm$0.54 & 21.21$\pm$1.18 & 38.56$\pm$0.83 &  35.21$\pm$0.42     &  53.56$\pm$0.75  \\
\hline
FedSeq & 81.81$\pm$1.37   &  82.56$\pm$0.53  &   54.66$\pm$2.85     &  65.40$\pm$1.09  &     43.22$\pm$0.61   &   52.49$\pm$0.33 & 33.01$\pm$0.70    &     62.88$\pm$5.64  \\
CWC & 81.85$\pm$1.38 & 82.94$\pm$0.37 & 55.16$\pm$1.52    & 65.63$\pm$1.33 &  44.61$\pm$1.40  & 52.27$\pm$0.58      & 32.28$\pm$0.88  &   59.06$\pm$3.20 \\
\textbf{SFedKD}  &    \boldmath{}\textbf{83.66$\pm$1.42}\unboldmath{}    & \boldmath{}\textbf{83.42$\pm$1.12}\unboldmath{} &    \boldmath{}\textbf{61.56$\pm$2.32}\unboldmath{}   & \boldmath{}\textbf{72.77$\pm$1.07}\unboldmath{} &   \boldmath{}\textbf{46.34$\pm$0.46}\unboldmath{}    & \boldmath{}\textbf{55.53$\pm$0.42}\unboldmath{} &  \boldmath{}\textbf{40.15$\pm$0.71}\unboldmath{} &   \boldmath{}\textbf{65.28$\pm$0.59}\unboldmath{} \\
\myhline
\end{tabular}%

    \label{tab:main_results}
    \vspace{-5pt}
\end{table*}%

\section{Experiments}
\subsection{Experimental Setup}
\noindent\textbf{Datasets and Models.} We conduct thorough experiments on five public benchmark datasets, including Fashion-MNIST~\cite{xiao2017fashion}, CIFAR-10~\cite{krizhevsky2009learning}, CINIC-10~\cite{darlow2018cinic}, CIFAR-100~\cite{krizhevsky2009learning}, and HAM10000~\cite{tschandl2018ham10000}. Fashion-MNIST comprises of 28 $\times$ 28 grayscale images of 70,000 fashion products. The CIFAR-10 dataset consists of 50,000 training images and 10,000 testing images, each with a size of 3 $\times$ 32 $\times$ 32. CINIC-10 is an extension of CIFAR-10 via the addition of downsampled ImageNet~\cite{deng2009imagenet} images. CIFAR-100 has the same format as CIFAR-10 with 100 classes of images. The HAM10000 dataset is an image dataset used for skin lesion classification in the medical field with 7 classes. We use a LeNet-5~\cite{lecun1998gradient} for Fashion-MNIST, an 8-layer AlexNet~\cite{krizhevsky2012imagenet} for CIFAR-10 and CINIC-10, and a simple CNN for other image datasets.

\noindent\textbf{Data Partition.} To simulate the heterogeneous data distribution, we use an extended Dirichlet strategy~\cite{li2024convergence} to partition the datasets. The extended Dirichlet strategy can be denoted by ExDir($C, \alpha$), which introduces an additional parameter $C$ to determine the number of classes per client compared with the Dirichlet distribution~\cite{wang2023delta}. Specifically, we randomly allocate $C$ different classes to each client and then draw $\mathbf{p}_c\sim \text{Dir}(\alpha \mathbf{q}_c)$ to allocate a proportion of the samples of class $c$ to each client. For example, $\mathbf{q}_c = [1, 1, 0, 0, \ldots,]$ means that the samples of class $c$ are only allocated to the first 2 clients. This strategy significantly enhances the flexibility of our design for federated learning in heterogeneous scenarios. For the Fashion-MNIST, CIFAR-10, and CINIC-10 datasets, we configure two settings: ExDir(2,0.5) and ExDir(2,10.0), while for CIFAR-100 and HAM10000, we implement ExDir(20,1.0) and ExDir(2,1.0), respectively. As the HAM10000 dataset has a lower class number than other datasets, we only set one case for it, and so does CIFAR-100 due to its large number of classes.


\noindent\textbf{Baselines.} We compare SFedKD with two SFL baselines: \textbf{FedSeq}~\cite{li2024convergence}, which is the most representative SFL algorithm; \textbf{CWC}~\cite{song2024cyclical}, a method based on cyclical weight consolidation to mitigate catastrophic forgetting in SFL. In addition, we compare our method with four PFL baselines: \textbf{FedAvg}~\cite{mcmahan2017communication}, which is the pioneering PFL method; \textbf{FedProx}~\cite{li2020federated}, a classic regularization-based PFL method; \textbf{FedCurv}~\cite{shoham2019overcoming}: a PFL method based on curvature adjustment regularization; \textbf{FedNTD}~\cite{lee2022preservation}, which preserves the global knowledge by not-true distillation in PFL.

\noindent\textbf{Evaluation Metrics.} We employ the Top-1 Accuracy to evaluate the model performance, which is the proportion of correctly classified test samples to the total number of test samples. Moreover, we use the time spent on local client training to evaluate the training efficiency, which is calculated by accumulating the time used in each training round.


\noindent\textbf{Implementation details.} The overall framework of SFedKD is implemented with Pytorch~\cite{paszke2019pytorch}. In our setting, we divide the dataset into 100 clients, then randomly sample 10 clients and select 5 teachers in each round. We run federated learning for 1000 rounds. The number of local epochs and batch size are 5 and 64, respectively. We use SGD optimizer with a 0.01 learning rate and the weight decay is set to 1e-4. To calculate the distance between data distributions, we employ KL-Divergence to measure the discrepancy.

\noindent\textbf{Experiment settings in Fig.~\ref{fig:fogetting}.} We implement the FedSeq~\cite{li2024convergence} algorithm to simulate the SFL process and split data using the Dirichlet distribution. The dataset is distributed among 10 clients, where all clients are selected in random order each round and train for 10 rounds, collectively forming a client sequence of length 100 (10 clients × 10 rounds). The i.i.d. and non-i.i.d. scenarios follow Dir($\alpha=100.0$) and Dir($\alpha=0.1$), respectively. All experiments are conducted on CIFAR-10 and CIFAR-100 using a CNN model. All other parameters follow the above Implementation details.

\subsection{Performance Analysis}
Table~\ref{tab:main_results} presents the test accuracy of our SFedKD and all baseline methods on five datasets under different data partitioning settings. The results indicate that our method outperforms all baselines across different datasets and heterogeneity levels. For instance, on the relatively simple FashionMNIST dataset, SFedKD achieves 83.66\% accuracy, at least a 1.81\% improvement over all baselines on Exdir(2,0.5) case. On CIFAR-10 and CINIC-10 under the ExDir(2,0.5) setting, it respectively achieves accuracy gains of more than 6.40\%, 1.73\% over other SFL methods and at least 29.08\%, 17.53\% over PFL methods. Moreover, at least 4.94\% and 2.40\% improvements are achieved on CIFAR-100 and HAM10000, respectively. Additionally, we observe that as data heterogeneity increases, the limitation of PFL methods becomes more pronounced, whereas our algorithm demonstrates a progressively greater advantage. This is because the PFL framework struggles to address extreme data heterogeneity, leading to a significant drop in model performance. Although these methods prevent local model drift by constraining local updates or preserving global knowledge, they fail to effectively extract knowledge across clients, resulting in poor generalization of the global model. Despite SFL can accelerate the convergence of the global model, the non-IID nature of client data exacerbates catastrophic forgetting. Our method leverages the strong convergence guarantees of the SFL framework in heterogeneous scenarios while employing knowledge distillation to maintain a balance between new and old knowledge, thereby achieving optimal performance.

\subsection{Ablation Study}

\subsubsection{Effects of discrepancy-aware multi-teacher knowledge distillation.} In our work, we extend the single-teacher Decoupled Knowledge Distillation approach to our multi-teacher setting and assign distinct weights to its two decoupled components. We conduct ablation experiments by replacing our KD method with multi-teacher KD frameworks AVER~\cite{you2017learning}, EBKD~\cite{kwon2020adaptive}, and CA-MKD~\cite{zhang2022confidence} to demonstrate the superiority of our KD method. In addition, we also conduct ablation experiments by replacing our teacher weighting strategy (i.e., two distinct weights) with average weights to verify its effectiveness. The CIFAR-10 dataset is partitioned using Exdir(2,0.5) and Exdir(2,10.0), and the results are shown in Table~\ref{tab:weights}. According to these results, our approach outperforms the method that enables other multi-teacher KD frameworks on FedSeq. Furthermore, both teacher weights positively contribute to our algorithm, and the best performance is achieved when both weights are enabled.

\begin{table}[t]
  \centering
  \renewcommand{\arraystretch}{1.1} 
  \caption{Ablation studies on our KD method.}
    \begin{tabular}{c|cc|cc}
    \myhline
    \textbf{Method}  & $g_k$  & $h_k$ & \textbf{Exdir(2,0.5)} & \textbf{Exdir(2,10.0)} \\
    \hline
    FedSeq & \textemdash   & \textemdash & 50.61\% & 63.50\% \\
    CWC &   \textemdash &\textemdash  & 52.75\% & 63.03\% \\
    FedSeq+AVER &   \textemdash & \textemdash & 60.76\% & 71.02\% \\
    FedSeq+EBKD & \textemdash   & \textemdash & 60.68\% & 71.28\% \\
    FedSeq+CA-MKD &  \textemdash  & \textemdash & 61.30\% & 71.58\% \\
    SFedKD &  $\times$  & $\times$ & 60.42\% & 70.92\% \\
    SFedKD & $\checkmark$ &  $\times$ & 60.75\% & 71.75\% \\
    SFedKD & $\times$  & $\checkmark$ & 63.89\% & 72.03\% \\
    SFedKD  & $\checkmark$ & $\checkmark$ & \textbf{64.33\%} & \textbf{72.30\%} \\
    \myhline
    \end{tabular}%
  \label{tab:weights}%
  \vspace{-5pt}
  
\end{table}%

\subsubsection{Effects of $\gamma$ and $\beta$.} 
\begin{figure}[t]
  \centering
  \begin{subfigure}{0.48\linewidth}
    \includegraphics[width=\linewidth]{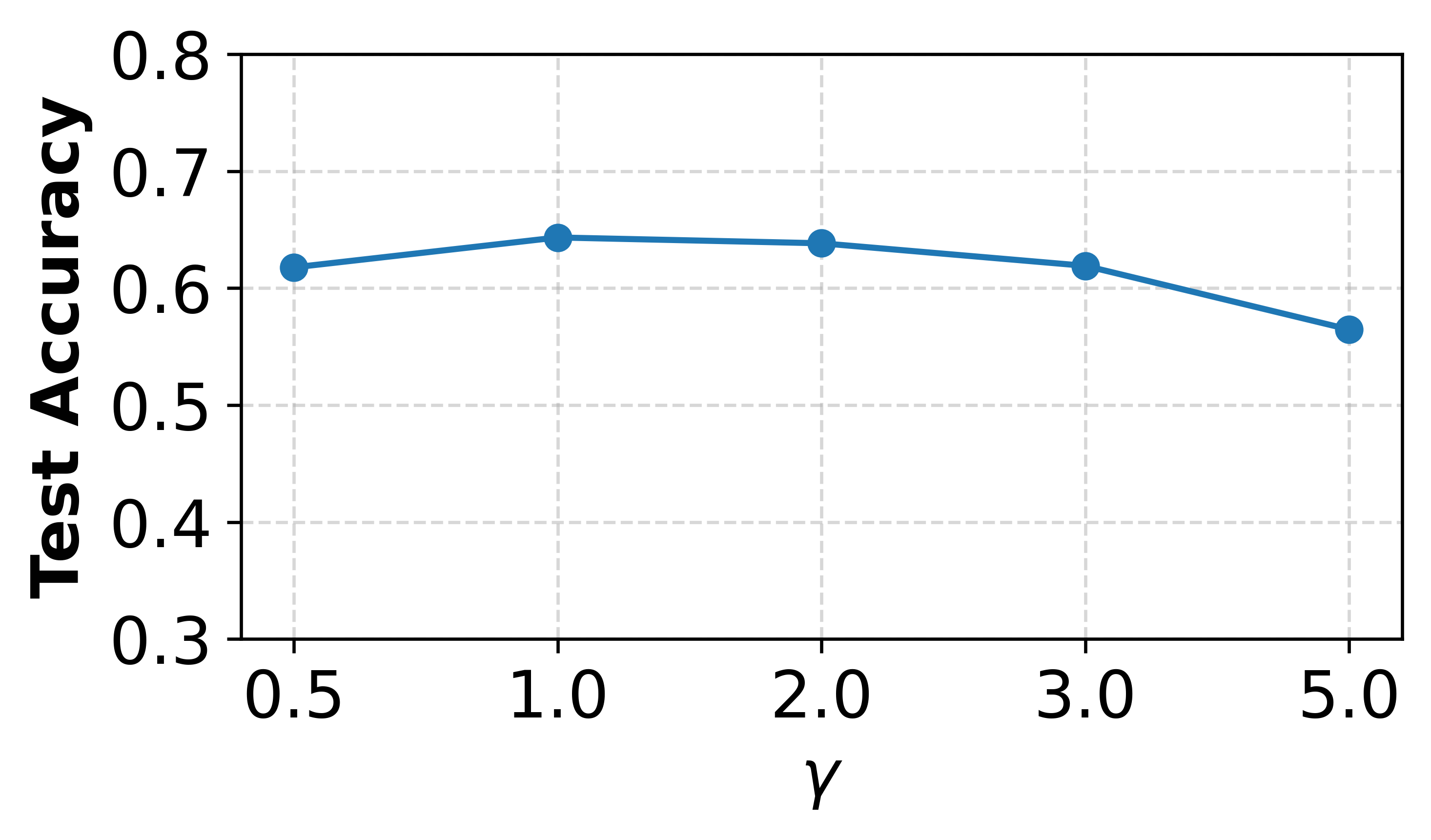}
    \caption{Impact of hyperparameter $\gamma$}
  \end{subfigure}
  \hspace{0.5mm}
  \begin{subfigure}{0.48\linewidth}
    \includegraphics[width=\linewidth]{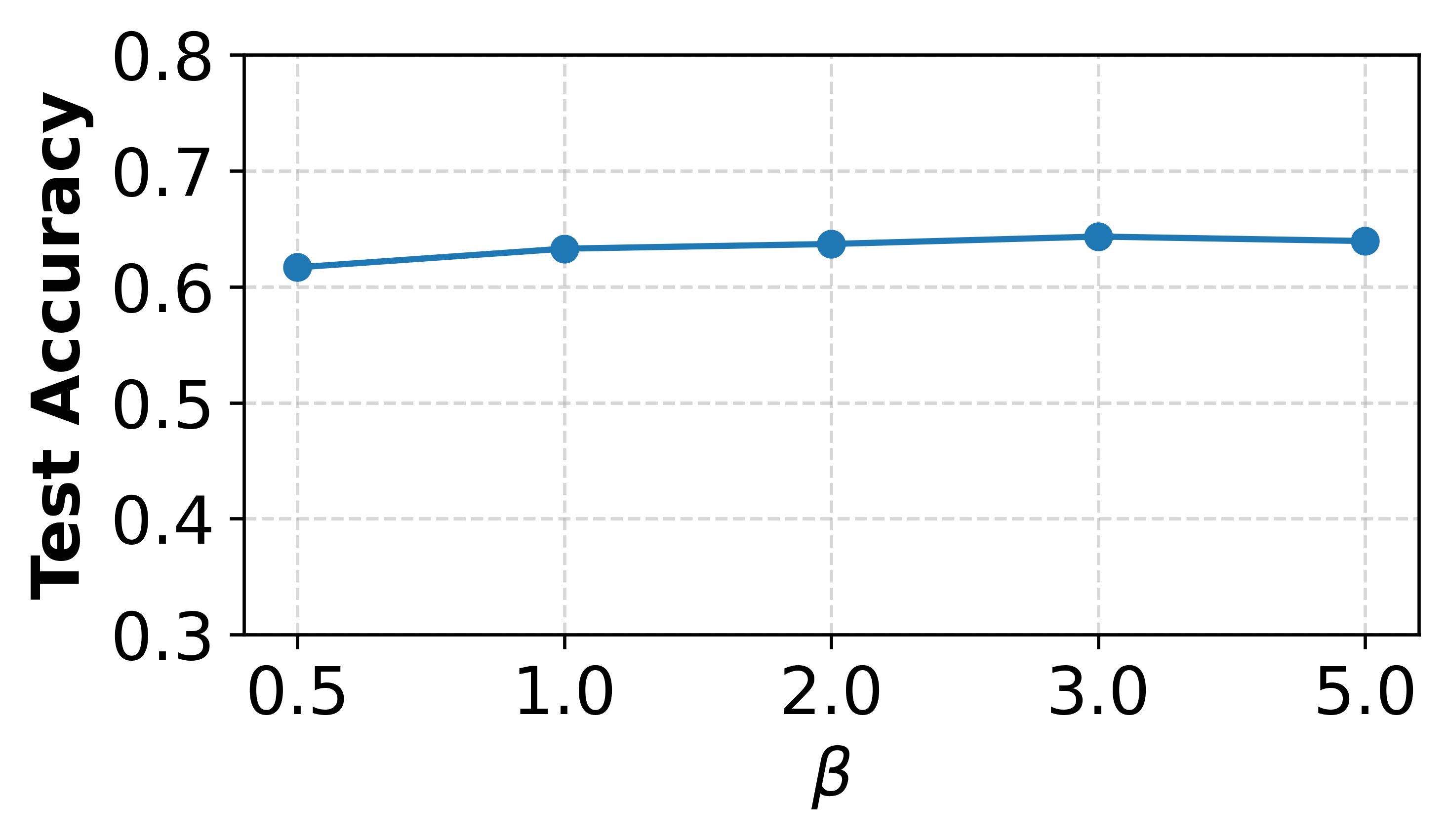}
    \caption{Impact of hyperparameter $\beta$}
  \end{subfigure}
  \vspace{-5pt}
\caption{The sensitivity analysis of $\gamma$ and $\beta$ on CIFAR-10.} 
   \label{fig:hyper}
   \vspace{-10pt}
\end{figure}

To investigate the impact of the hyperparameters $\gamma$ and $\beta$ on the experimental results, we conduct experiments on the CIFAR-10 dataset with data heterogeneity set to Exdir(2,0.5), as shown in Fig.~\ref{fig:hyper}. We vary $\gamma$ and $\beta$ within the range $\{0.5, 1.0, 2.0, 3.0, 5.0\}$. Specifically, we fix $\beta=3.0$ and change $\gamma$, followed by fixing $\gamma=1.0$ and varying $\beta$. We observe that the overall experimental results exhibit robustness in a wide range of hyperparameters. Additionally, the optimal parameter combination ($\gamma=1.0$, $\beta=3.0$) was adopted in the main experiments.

\begin{figure}[t]
  \centering
  \begin{subfigure}{0.48\linewidth}
    \includegraphics[width=\linewidth]{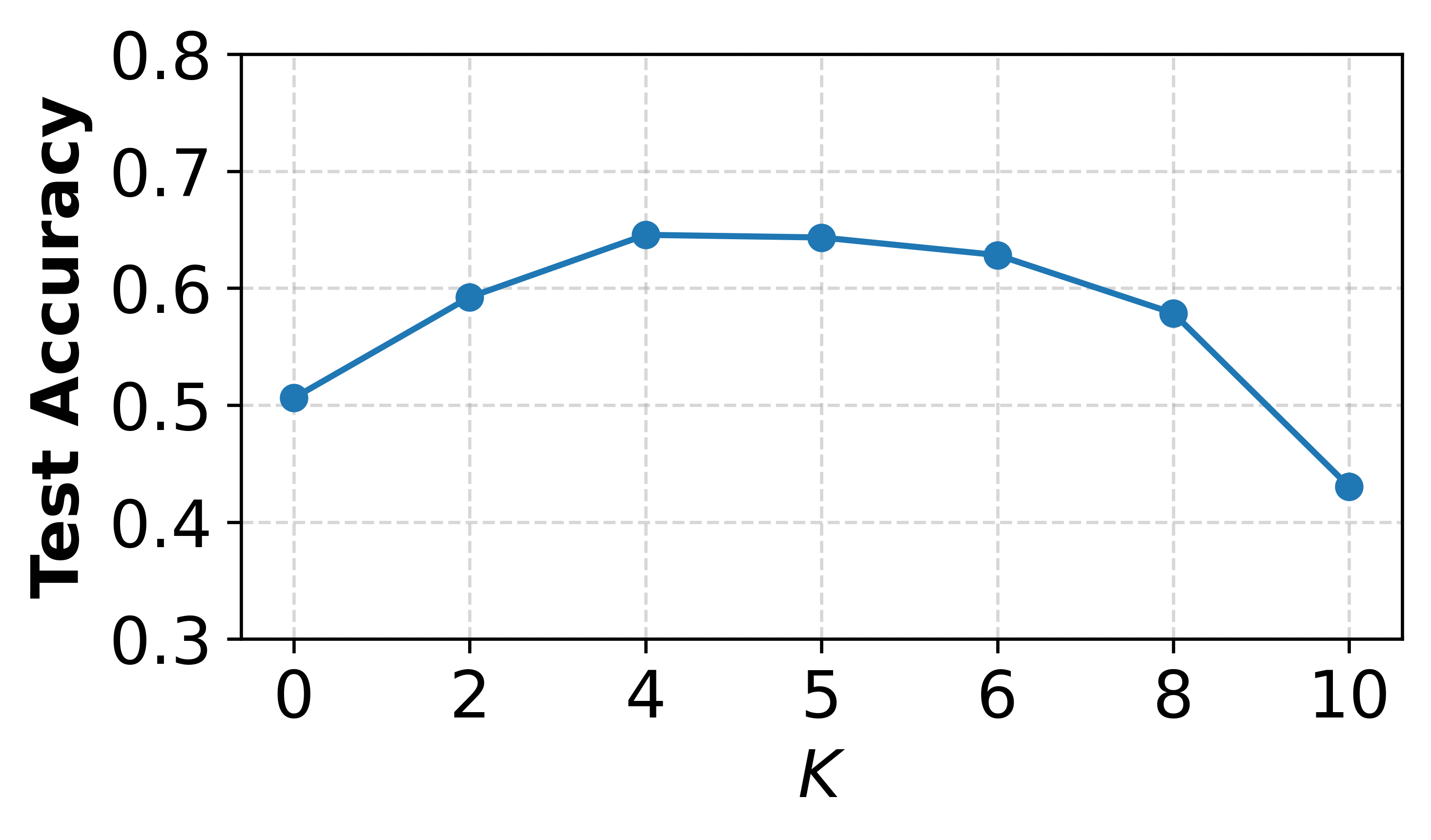}
    \caption{Exdir(2,0.5)}
  \end{subfigure}
  \hspace{0.5mm}
  \begin{subfigure}{0.48\linewidth}
    \includegraphics[width=\linewidth]{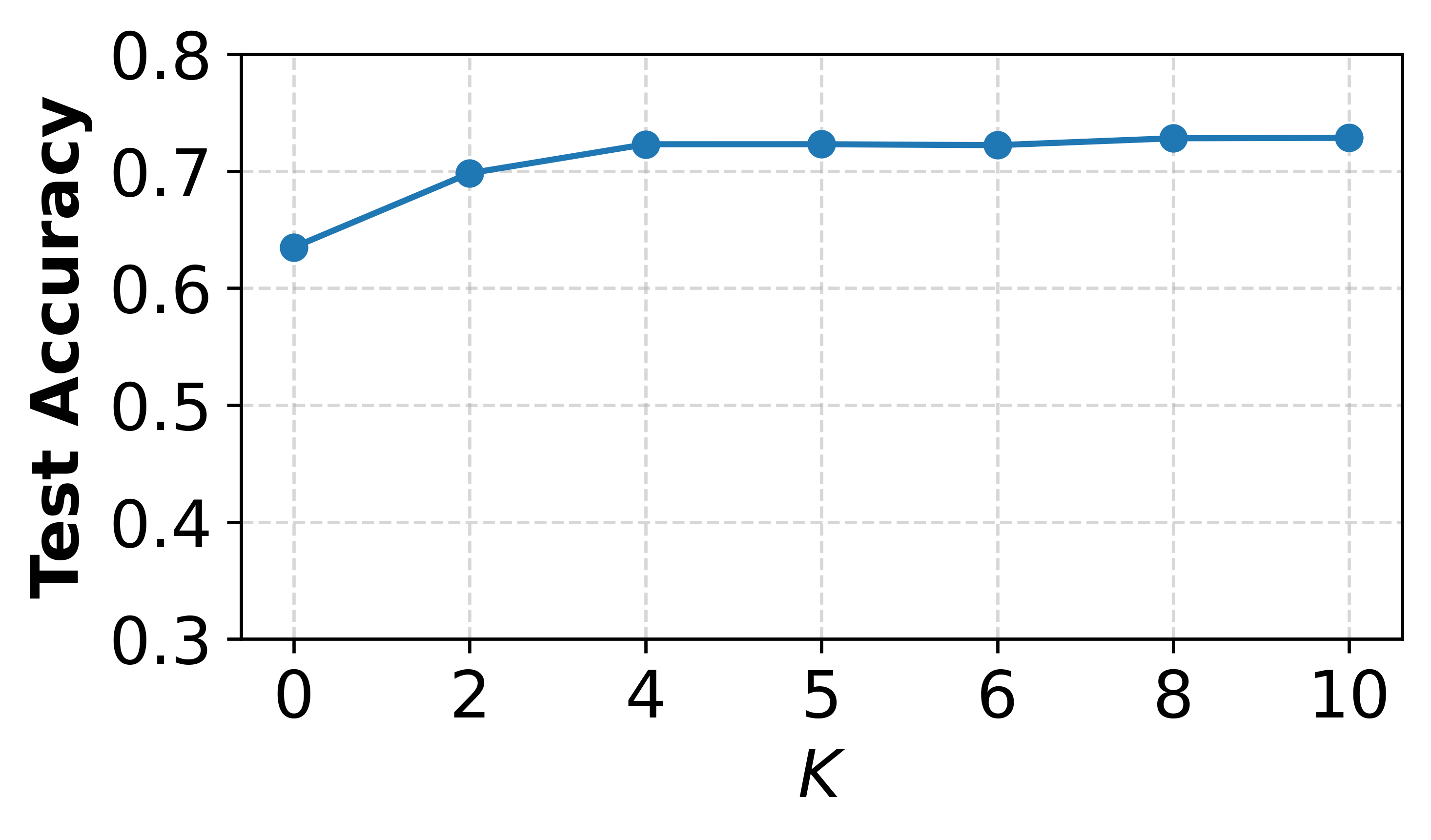}
    \caption{Exdir(2,10.0)}
  \end{subfigure}
  \vspace{-5pt}
\caption{The impact of teacher number $K$ on CIFAR-10.} 
   \label{tab:teacher}
   \vspace{-10pt}
\end{figure}




\begin{figure*}[t]
  \centering
  \begin{subfigure}{0.24\linewidth}
    \includegraphics[width=\linewidth]{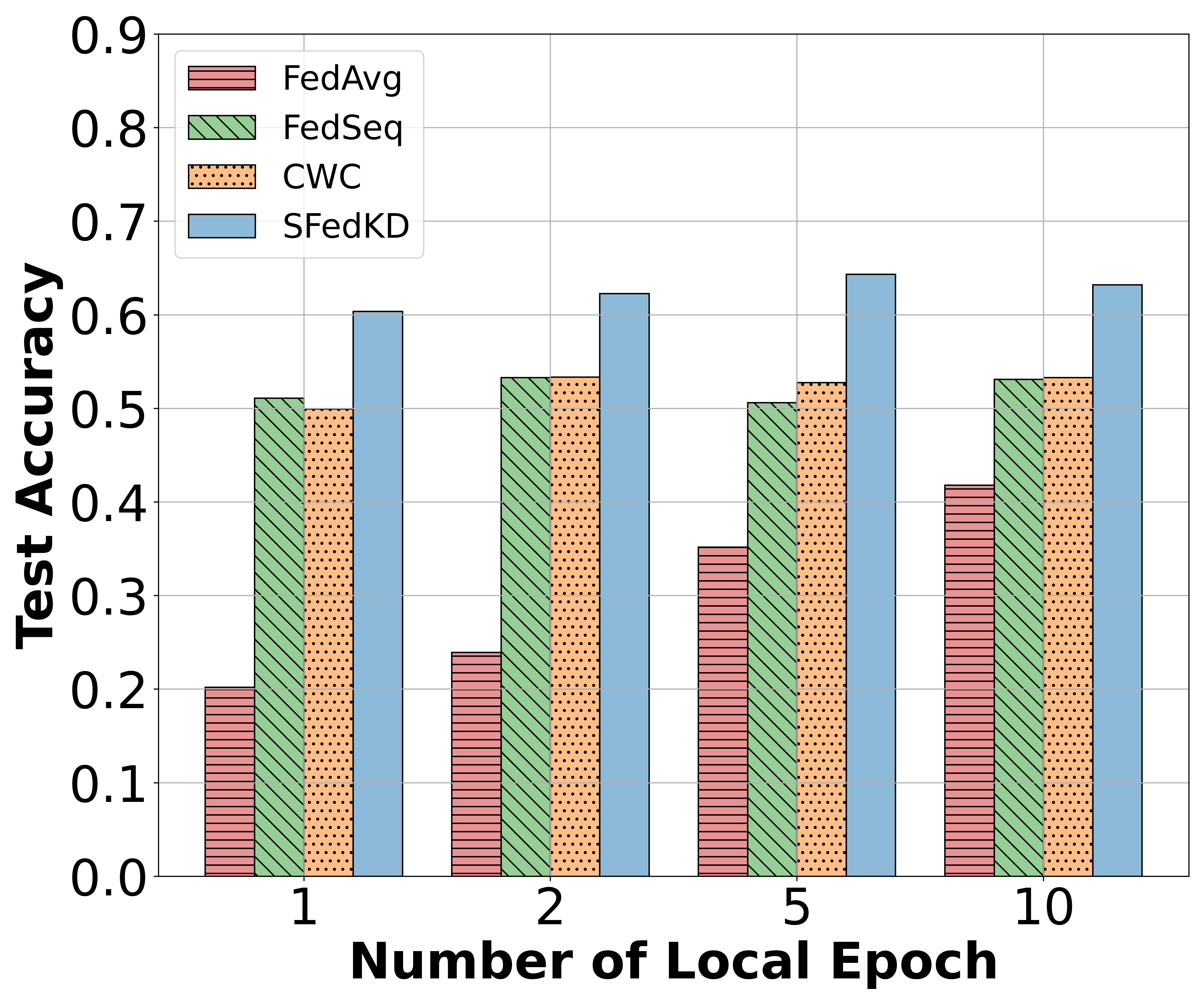}
    \caption{Local Epoch}
    \label{fig:P}
  \end{subfigure}
  \hspace{0.5mm}
  \begin{subfigure}{0.24\linewidth}
    \includegraphics[width=\linewidth]{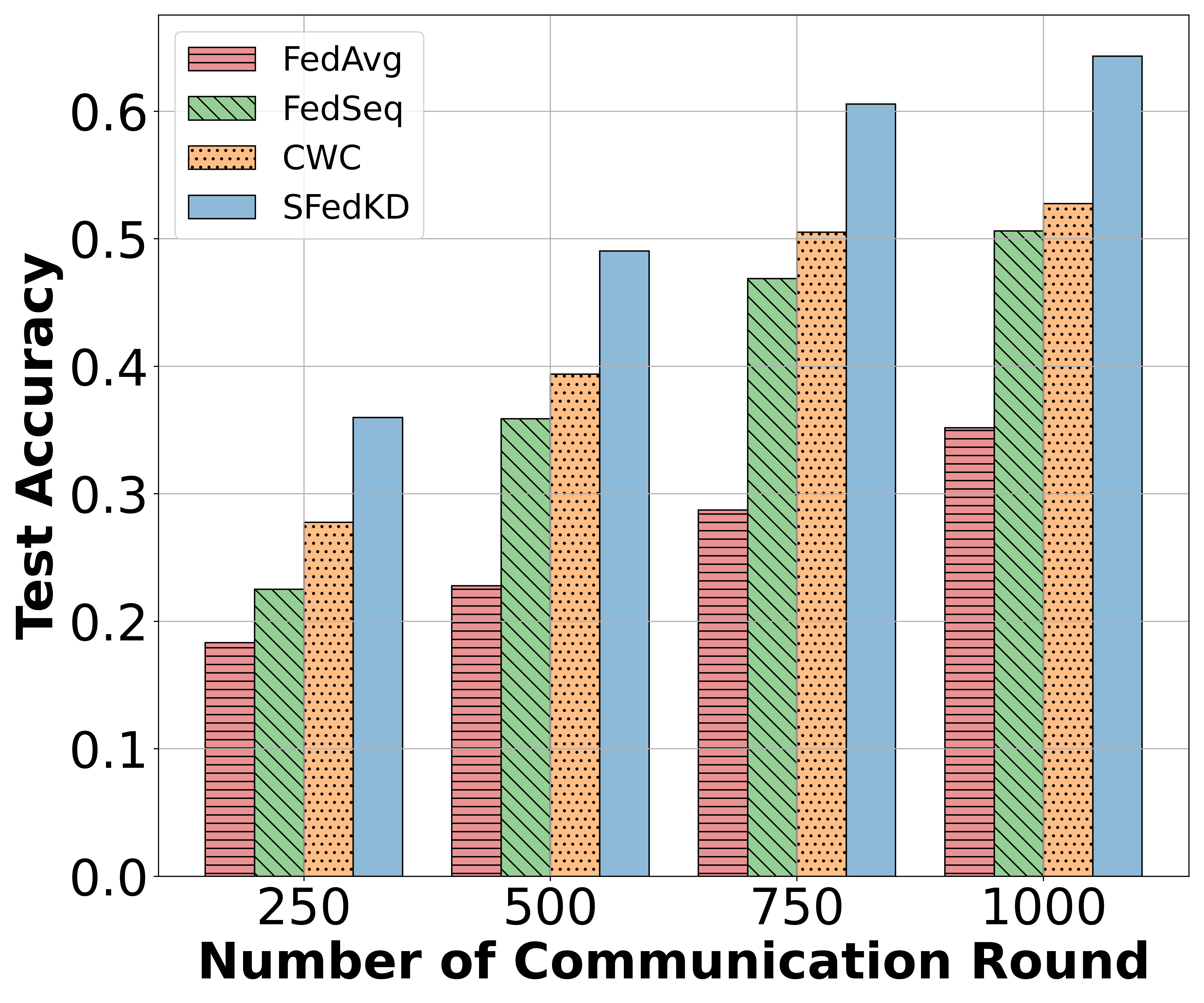}
    \caption{Training Round}
    \label{fig:R}
  \end{subfigure}
  \hspace{0.5mm}
  \begin{subfigure}{0.24\linewidth}
    \includegraphics[width=\linewidth]{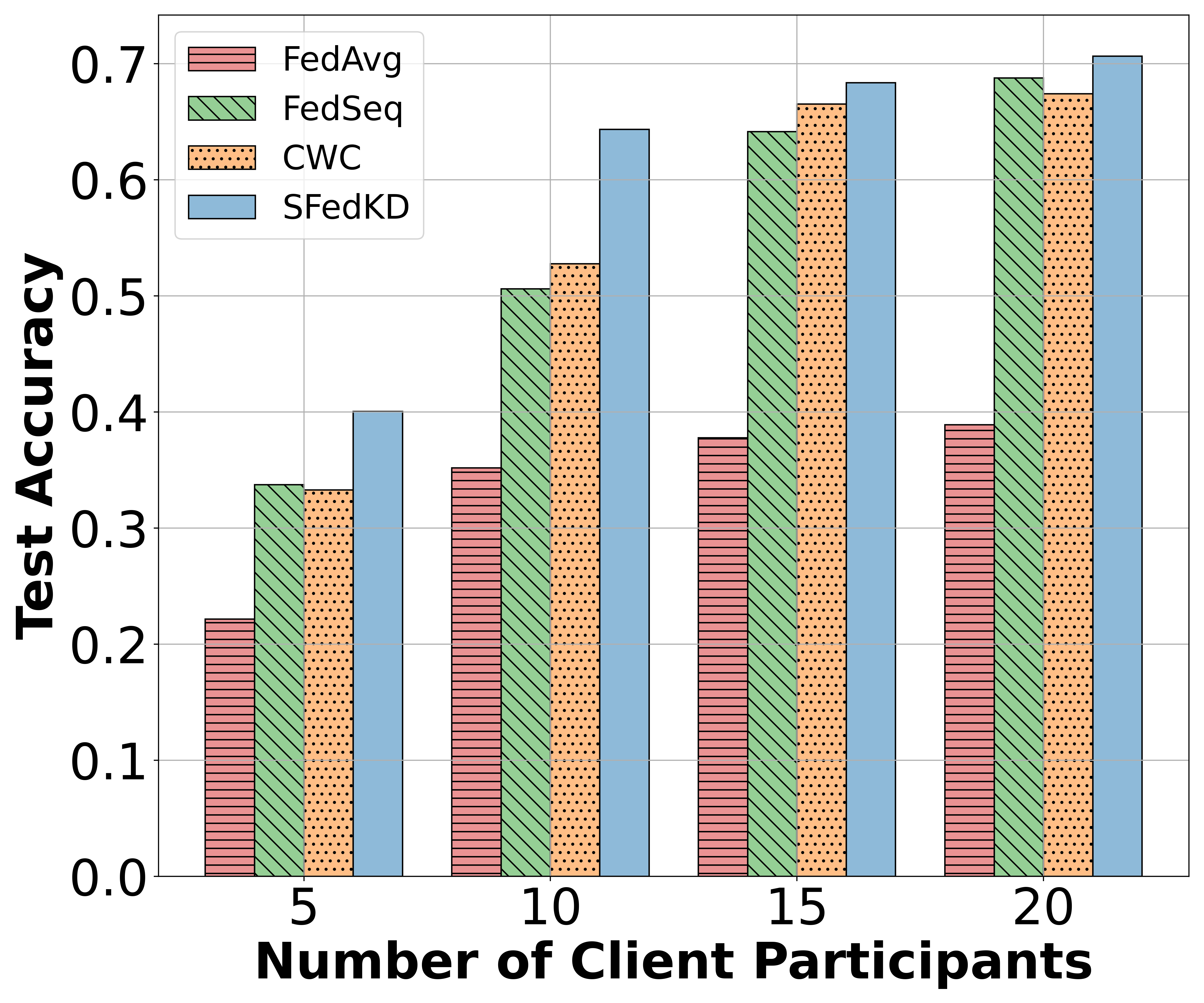}
    \caption{Number of Clients}
    \label{fig:noniid}
  \end{subfigure}
  \hspace{0.5mm}
  \begin{subfigure}{0.24\linewidth}
    \includegraphics[width=\linewidth]{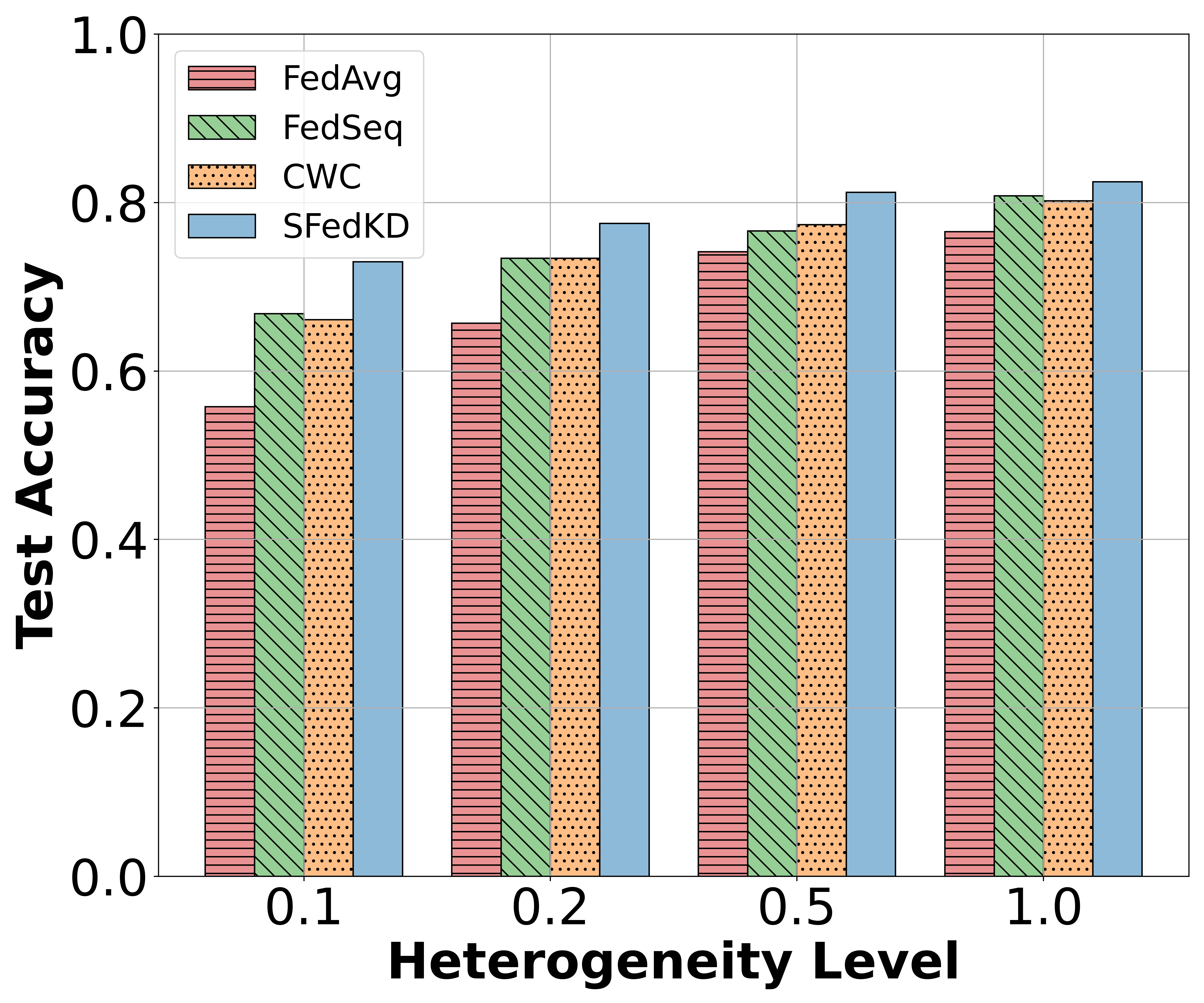}
    \caption{Heterogeneity Level}
    \label{fig:K}
  \end{subfigure}
  \vspace{-5pt}
\caption{The impact of four key FL arguments on performance.} 
   \label{fig:argument}
\end{figure*}

\begin{table}[h]
    \centering
    \renewcommand{\arraystretch}{1.1}  
    \caption{Performance comparison of our teacher selection algorithm and random sampling.}
    \resizebox{\linewidth}{!}{
    \begin{tabular}{c|c|cccc}
        \myhline
         & $K$ & 3 & 4 & 5 & 6 \\
        \hline
        \multirow{2}{*}{Exdir(2,0.5)}
        & Random  & 58.14\%  & 57.55\%  & 59.01\%  & 61.92\%  \\
        & TS  & \textbf{59.70\%}  & \textbf{61.93\%}  & \textbf{62.34\%}  & \textbf{62.83\%}  \\
        \hline
        & Advantage &  1.56\%  & 4.38\%  & 3.33\%  & 0.91\%  \\
        \hline
        \multirow{2}{*}{Exdir(2,1.0)}
        & Random  & 63.24\%  & 62.92\%  & 65.27\%  & 65.13\%  \\
        & TS  & \textbf{63.45\%}  & \textbf{64.92\%}  & \textbf{66.09\%}  & \textbf{65.91\%}  \\
        \hline
        & Advantage  & 0.21\%  & 2.00\%  & 0.82\% & 0.78\%  \\
        \hline
        \multirow{2}{*}{Exdir(2,10.0)}
        & Random  & 70.90\%  & 71.17\%  & 70.82\%  & 72.10\%  \\
        & TS  & \textbf{71.03\%}  & \textbf{71.90\%}  & \textbf{71.99\%}  & \textbf{72.22\%}  \\
        \hline
        & Advantage  & 0.13\%  & 0.73\%  & 1.17\%  & 0.12\%  \\
        \hline
        \multirow{2}{*}{Exdir(2,100.0)}
        & Random  & 69.59\%  & 69.53\%  & 70.56\%  & 70.16\%  \\
        & TS  & \textbf{69.88\%}  & \textbf{70.96\%}  & \textbf{71.30\%}  & \textbf{70.45\%}  \\
        \hline
        & Advantage  & 0.29\%  & 1.43\% & 0.74\%  & 0.29\% \\
        \myhline
    \end{tabular}}
    \label{tab:ts}
    \vspace{-5pt}
\end{table}


\subsubsection{Effects of teacher number.} We set different teacher numbers $K$ on CIFAR-10 Exdir(2,0.5) and Exdir(2,10.0) cases, as depicted in Fig.~\ref{tab:teacher}. When $K=0$, the setup is equivalent to the FedSeq algorithm. We observe that once the number of teachers exceeds a certain threshold, the model's performance ceases to improve and may even degrade. This further validates our hypothesis that redundant teacher models exist and highlights the necessity of our proposed teacher selection mechanism.


\subsubsection{Effects of teacher selection mechanism.} We conduct a series of comparative experiments on the CIFAR-10 dataset, comparing our Teacher Selection (TS) algorithm with random sampling. As shown in Table~\ref{tab:ts}, under four data partitioning scenarios, when the number of selected teachers $K\in \{3,4,5,6\}$, our proposed strategy consistently outperforms random sampling. This result confirms that our algorithm achieves more comprehensive knowledge space coverage. Notably, our method shows greater advantages in more heterogeneous environments, as random sampling will achieve complementary teacher selection results with lower probabilities.

\subsubsection{Effects of different choices of parameters.}  We tune four key arguments in FL, including local epoch ($E\in\{1, 2, 5, 10\}$), training round ($R\in\{250, 500, 750, 1000\}$), the number of selected clients ($M\in\{5, 10, 15, 20\}$), and heterogeneity level ($\alpha \in \{0.1,0.2,0.5,1.0\}$), and show the accuracy of our method and three baselines in Fig.~\ref{fig:argument}. Here, $\alpha$ is the concentration parameter of Dirichlet distribution $Dir(\alpha\mathbf{p})$, where $\mathbf{p}$ characterizes a prior distribution. A smaller $\alpha$ value indicates a higher heterogeneity level. The experimental results demonstrate that our method outperforms all baselines across different FL arguments. All the experiments in this part are conducted on the CIFAR-10 dataset.

\begin{table}[t]
\caption{Performance under different discrepancy metrics.}
\renewcommand{\arraystretch}{1.1}
\label{table:metric}
\begin{center}
\begin{tabular}{c|cccc}
\myhline
\textbf{Metric} & \textbf{L1} & \textbf{L2} & \textbf{JS} & \textbf{KL} \\
\myhline
Exdir(2,0.5) & 63.31\% & 63.10\% & 63.90\% & 64.33\% \\
\myhline
Exdir(2,10.0) & 72.78\% & 72.10\% & 72.42\% & 72.30\% \\
\myhline
\end{tabular}
\end{center}
\vspace{-5pt}
\end{table}

\subsubsection{Effects of different discrepancy metrics.} To investigate the impact of different distribution discrepancy metrics, we compare the effectiveness of four metrics—L1 distance, L2 distance, KL-Divergence, and JS-Divergence~\cite{fuglede2004jensen} on the CIFAR-10 dataset under two data partitioning settings: Exdir(2,0.5) and Exdir(2,10.0). Table~\ref{table:metric} shows that our method achieves comparable performance across different metrics, demonstrating its robustness to the choice of discrepancy metrics. Throughout this paper, we adopt KL-Divergence for all experiments. 

\subsection{Efficiency Analysis}
We propose the Teacher Selection (TS) mechanism to enable efficient multi-teacher knowledge distillation. To demonstrate that TS can reduce training consumption, we conduct experiments on Fashion-MNIST, CIFAR-10, and CINIC-10 datasets under the ExDir(2,0.5) setting. We set multiple teacher numbers and record their training times separately. To ensure fairness, all experiments are conducted on an Intel(R) Xeon(R) Platinum 8352V CPU and an NVIDIA A40 (48GB) GPU, and 256GB RAM.

\begin{figure}[t]
  \centering
  \begin{subfigure}{0.31\linewidth}
    \includegraphics[width=\linewidth]{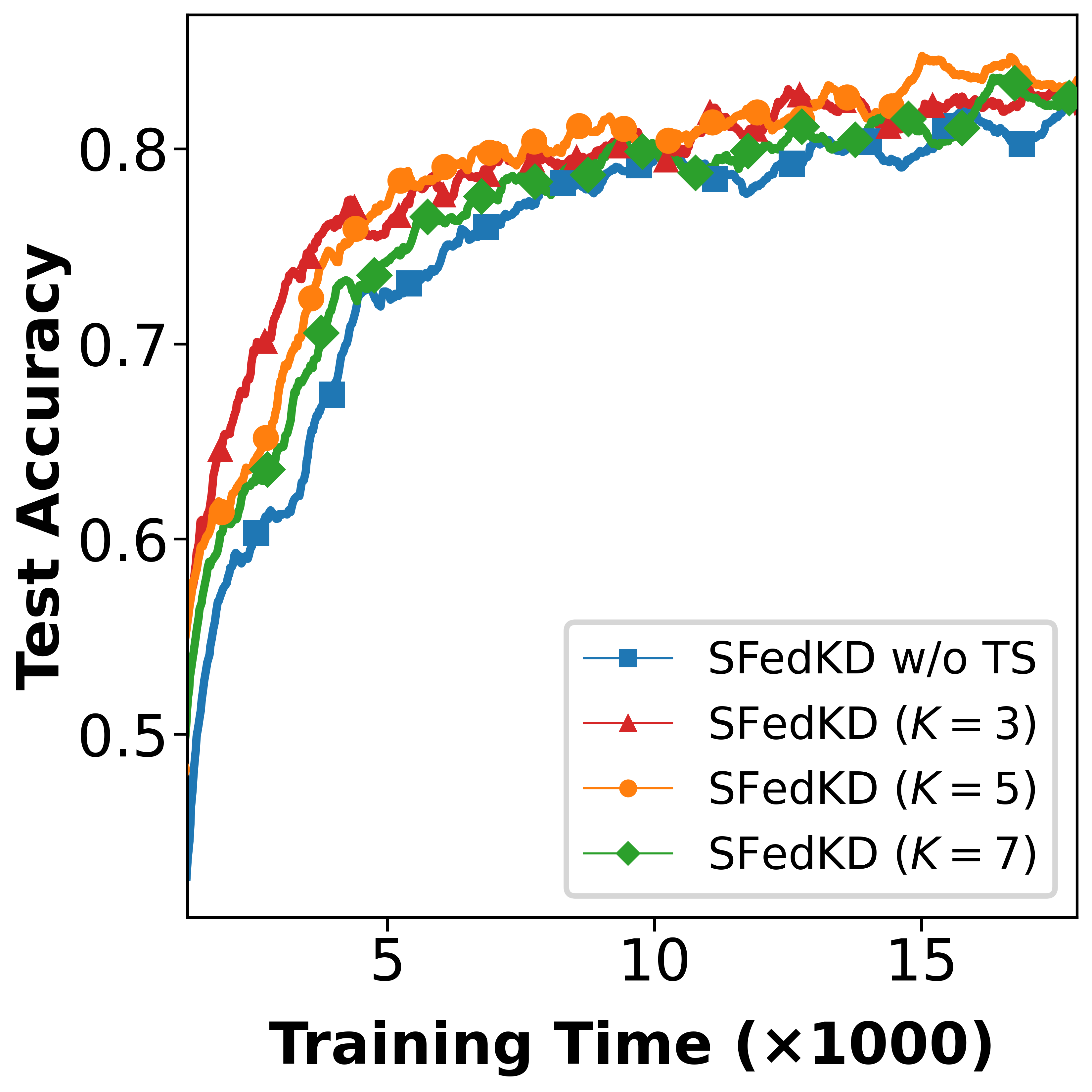}
    \caption{Fashion-MNIST}
  \end{subfigure}
  \hspace{0.5mm}
   \begin{subfigure}{0.31\linewidth}
    \includegraphics[width=\linewidth]{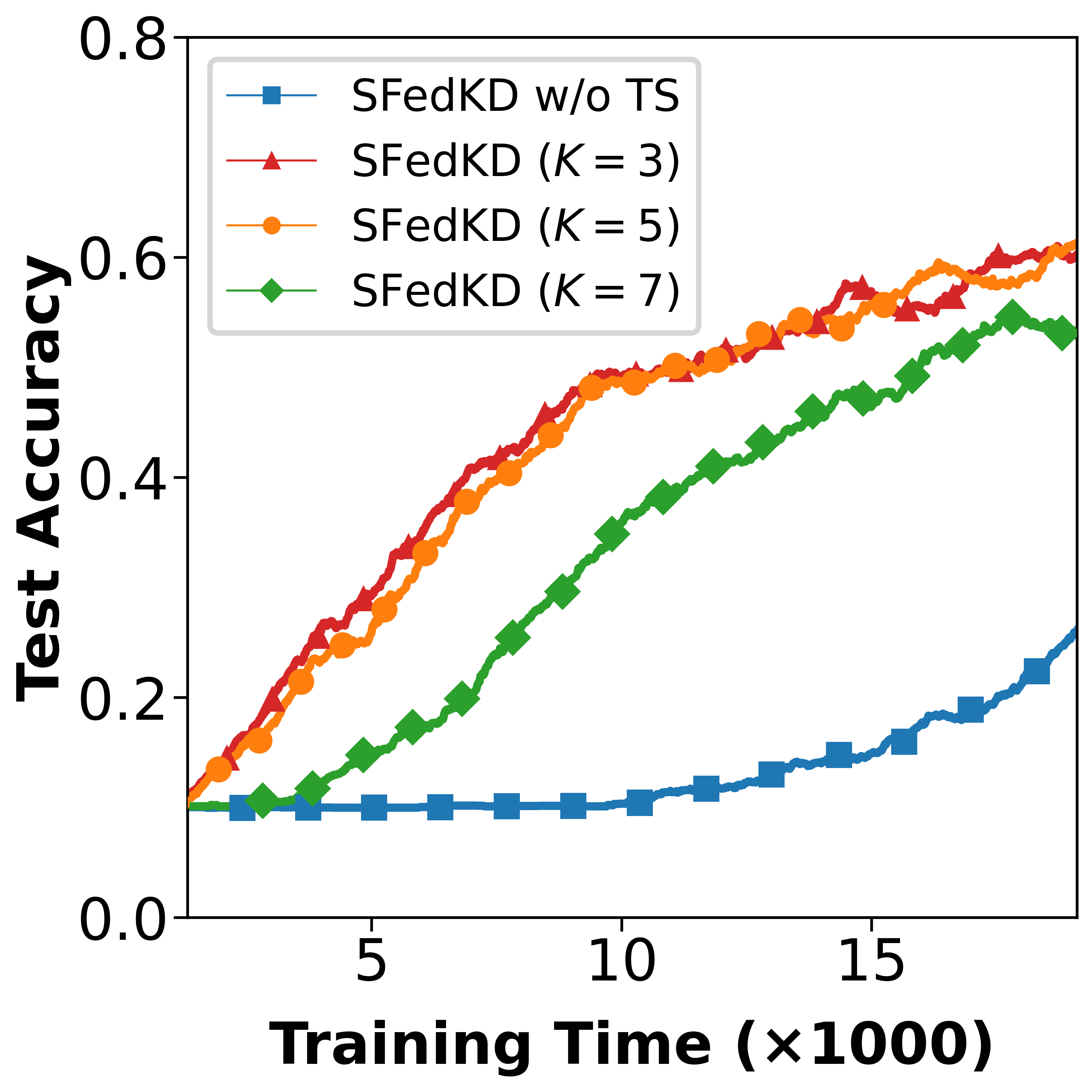}
    \caption{CIFAR-10}
  \end{subfigure}
  \hspace{0.5mm}
  \begin{subfigure}{0.31\linewidth}
    \includegraphics[width=\linewidth]{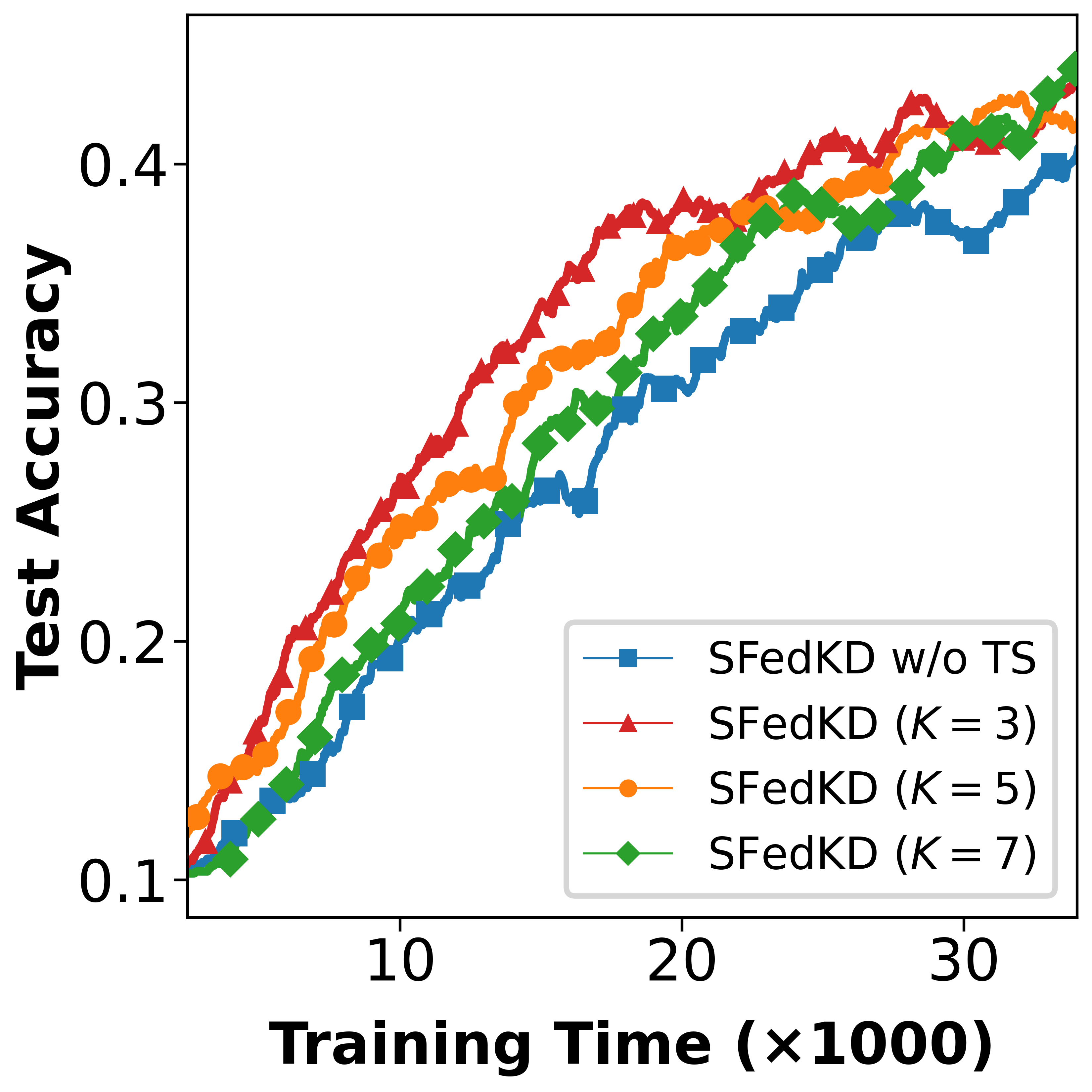}
    \caption{CINIC-10}
  \end{subfigure}
  \vspace{-5pt}
   \caption{Test accuracy over total training time.}
   \label{fig:acc}
   \vspace{-10pt}
\end{figure}

\begin{figure}[t]
  \centering
  \begin{subfigure}{0.31\linewidth}
    \includegraphics[width=\linewidth]{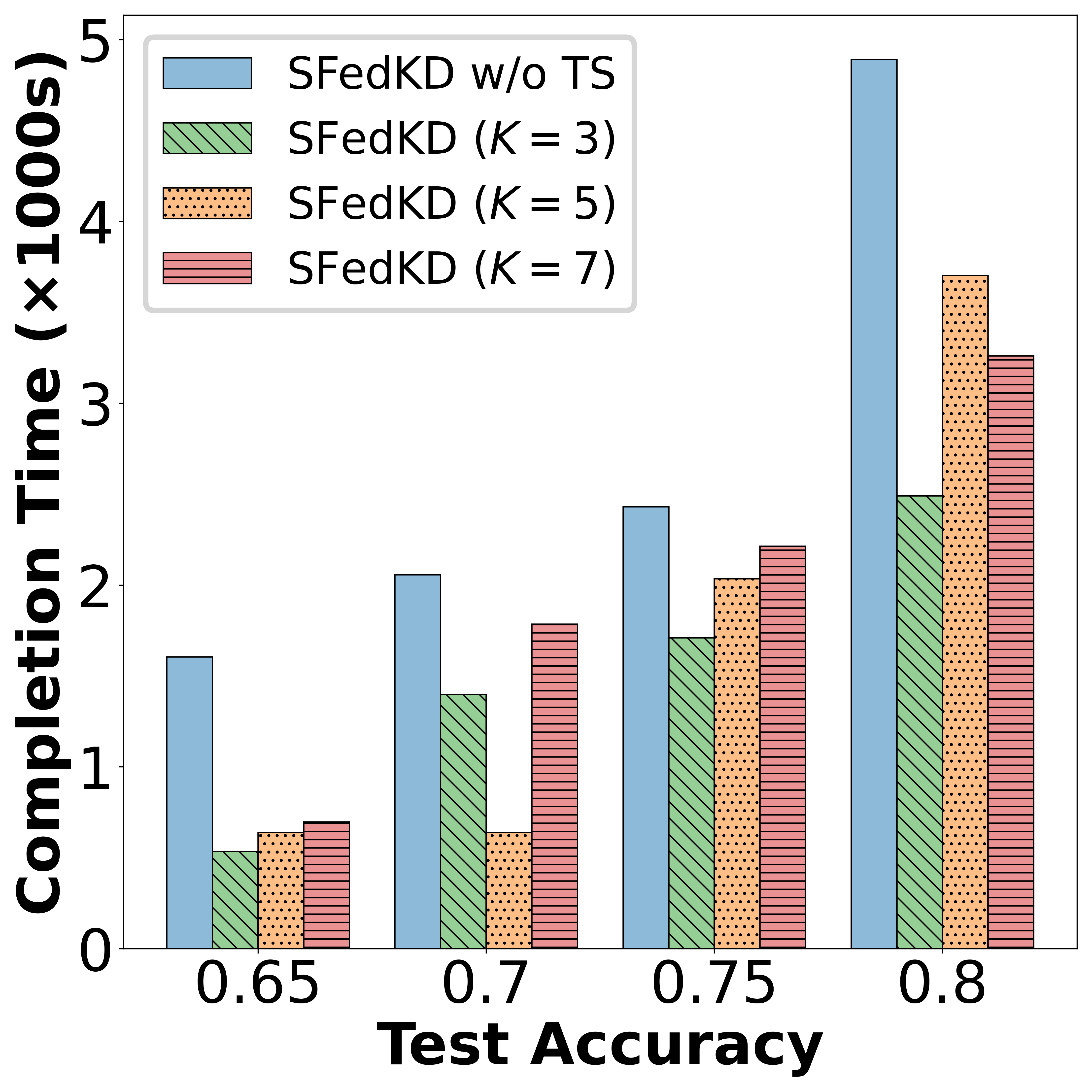}
    \caption{Fashion-MNIST}
  \end{subfigure}
  \hspace{0.5mm}
   \begin{subfigure}{0.31\linewidth}
    \includegraphics[width=\linewidth]{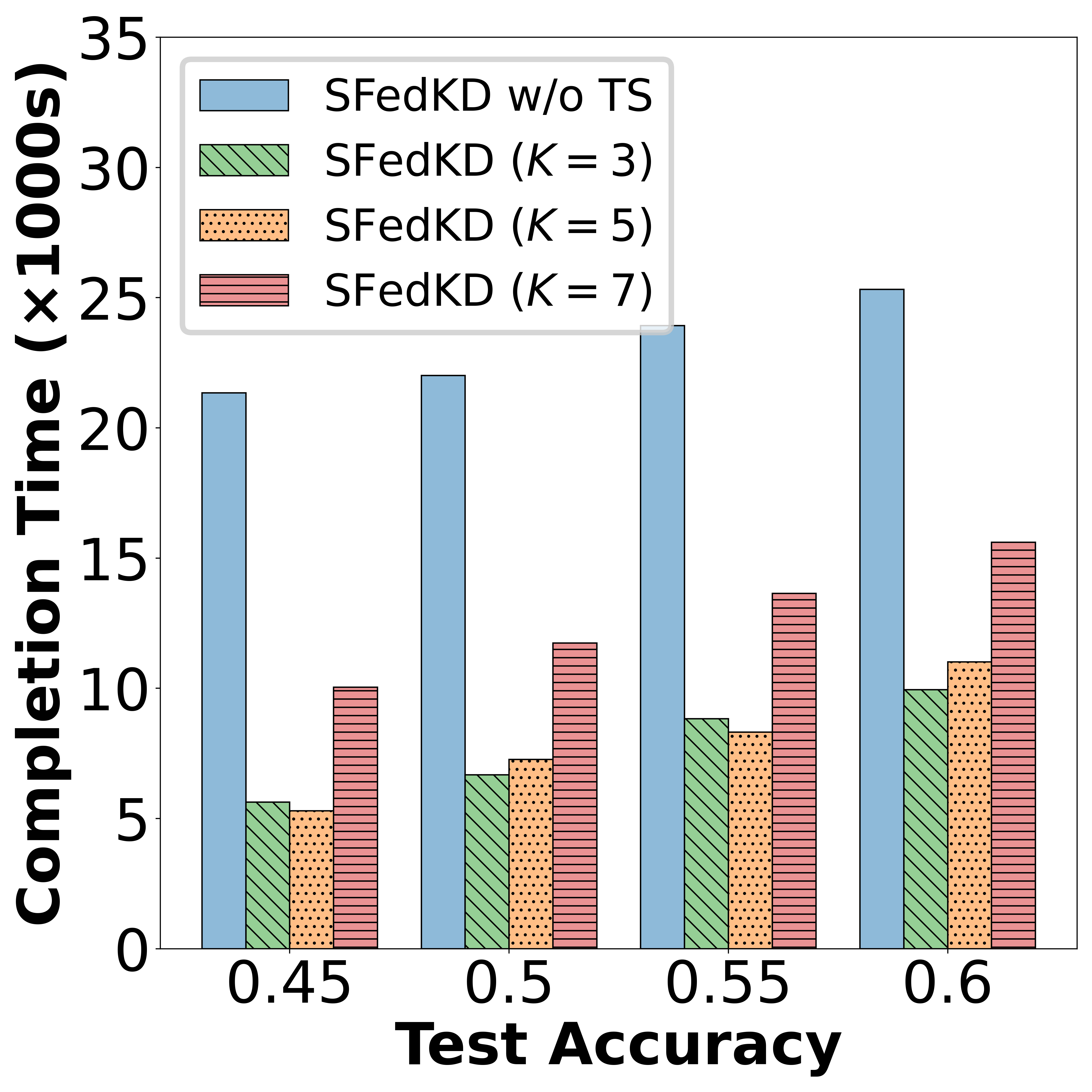}
    \caption{CIFAR-10}
  \end{subfigure}
  \hspace{0.5mm}
  \begin{subfigure}{0.31\linewidth}
    \includegraphics[width=\linewidth]{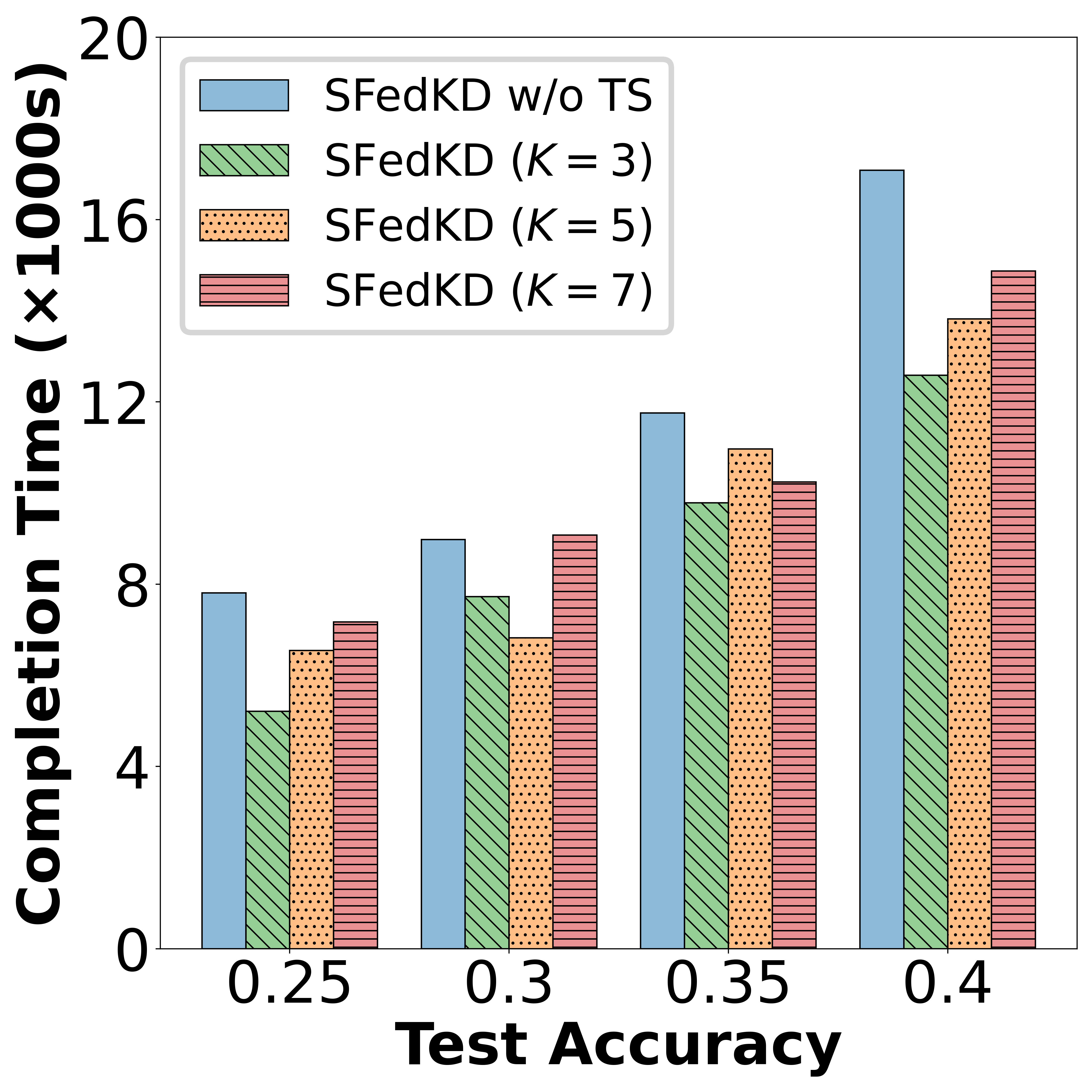}
    \caption{CINIC-10}
  \end{subfigure}
   \vspace{-5pt}
   \caption{Completion time under different target accuracies.}
   \label{fig:time}
   \vspace{-10pt}
\end{figure}

Fig.~\ref{fig:acc} shows the test accuracy under the same total training time when TS is disabled ($K=10$) and when the number of selected teachers $K$ is set to 3, 5, and 7. Fig.~\ref{fig:time} illustrates the completion time required for SFedKD to reach the target accuracies. Clearly, SFedKD achieves faster convergence speed when employing the TS mechanism. Moreover, for all target accuracy levels, selecting a subset of teachers consistently requires less time compared to using all teacher models. For instance, when $K \in \{3,5,7\}$, TS can separately speed up training to around 80\% accuracy by about 2.0$\times$, 1.3$\times$, and 1.5$\times$ on Fashion-MNIST, and reach the target accuracy of 60\% at 2.5$\times$, 2.3$\times$, and 1.6$\times$ speed on CIFAR-10. This further demonstrates that our TS mechanism enhances training efficiency while maintaining model performance. 

\section{Conclusion}
In this paper, we investigate the catastrophic forgetting problem in SFL caused by data heterogeneity. We propose SFedKD, an SFL framework with discrepancy-aware multi-teacher knowledge distillation. We extend Decoupled Knowledge Distillation to our multi-teacher setting by developing discrepancy-aware weighting schemes that assign distinct weights to teachers' target-class and non-target-class knowledge based on class distributional discrepancies with student data. Through personalized knowledge extraction from historical client models, our method effectively balances the learning of new and old knowledge, preserving global model performance on heterogeneous data. Furthermore, we introduce a complementary-based teacher selection mechanism that strategically selects teachers to minimize redundancy while maintaining knowledge diversity, which significantly enhances distillation efficiency. Experimental results validate the effectiveness of our approach, demonstrating improved model retention of past knowledge and overall performance in heterogeneous FL scenarios.

\bibliographystyle{ACM-Reference-Format}
\bibliography{sample-base}

\end{document}